\PassOptionsToPackage{unicode}{hyperref}
\PassOptionsToPackage{hyphens}{url}
\PassOptionsToPackage{dvipsnames,svgnames,x11names}{xcolor}

\documentclass[11pt]{article}

\usepackage[letterpaper,margin=1in]{geometry}
\usepackage{indentfirst}
\usepackage[T1]{fontenc}
\usepackage[utf8]{inputenc}
\usepackage{textcomp}
\usepackage{lmodern}

\DeclareUnicodeCharacter{201C}{``}
\DeclareUnicodeCharacter{201D}{''}

\usepackage{amsmath,amssymb,amsfonts,amsthm,mathtools}
\allowdisplaybreaks

\usepackage{graphicx}
\usepackage{booktabs}
\usepackage{array}
\usepackage{longtable}
\usepackage{caption}
\usepackage{subcaption}
\usepackage{float}

\makeatletter
\def\maxwidth{%
  \ifdim\Gin@nat@width>\linewidth
    \linewidth
  \else
    \Gin@nat@width
  \fi
}
\def\maxheight{%
  \ifdim\Gin@nat@height>\textheight
    \textheight
  \else
    \Gin@nat@height
  \fi
}
\setkeys{Gin}{width=\maxwidth,height=\maxheight,keepaspectratio}
\def\fps@figure{htbp}
\makeatother

\usepackage{algorithm}
\usepackage{algorithmicx}
\usepackage{algpseudocode}

\algrenewcommand\algorithmicrequire{\textbf{Input:}}
\algrenewcommand\algorithmicensure{\textbf{Output:}}
\algrenewcommand\algorithmicfor{\textbf{for}}
\algrenewcommand\algorithmicdo{\textbf{do}}
\algrenewcommand\algorithmicend{\textbf{end}}
\algrenewcommand\algorithmicif{\textbf{if}}
\algrenewcommand\algorithmicthen{\textbf{then}}
\algrenewcommand\algorithmicelse{\textbf{else}}
\algrenewcommand\algorithmicprocedure{\textbf{procedure}}

\usepackage{enumitem}

\makeatletter
\ifx\paragraph\undefined\else
  \let\oldparagraph\paragraph
  \renewcommand{\paragraph}{%
    \@ifstar
      \xxxParagraphStar
      \xxxParagraphNoStar
  }
  \newcommand{\xxxParagraphStar}[1]{\oldparagraph*{#1}\mbox{}}
  \newcommand{\xxxParagraphNoStar}[1]{\oldparagraph{#1}\mbox{}}
\fi
\ifx\subparagraph\undefined\else
  \let\oldsubparagraph\subparagraph
  \renewcommand{\subparagraph}{%
    \@ifstar
      \xxxSubParagraphStar
      \xxxSubParagraphNoStar
  }
  \newcommand{\xxxSubParagraphStar}[1]{\oldsubparagraph*{#1}\mbox{}}
  \newcommand{\xxxSubParagraphNoStar}[1]{\oldsubparagraph{#1}\mbox{}}
\fi
\makeatother

\newtheorem{theorem}{Theorem}
\newtheorem{definition}{Definition}

\newtheorem{note}{Remark}
\newtheorem{proposition}{Proposition}
\newtheorem{example}{Example}
\newtheorem{assumption}{Assumption}
\newtheorem{corollary}{Corollary}

\usepackage{csquotes}

\usepackage[colorlinks=true, linkcolor=blue, citecolor=blue, urlcolor=red]{hyperref}

\usepackage{xcolor}
\IfFileExists{xurl.sty}{\usepackage{xurl}}{}
\usepackage[
  backend=biber,
  style=authoryear,
  natbib = true,
  giveninits=true,
  maxcitenames=3,
  mincitenames=1,
  maxbibnames=99,
  uniquename=false,
  isbn=false
]{biblatex}

\renewbibmacro*{cite}{%
  \printtext[bibhyperref]{%
    \iffieldundef{shorthand}
      {\ifthenelse{\ifnameundef{labelname}\OR\iffieldundef{labelyear}}
        {\iffieldundef{label}
           {\printfield[citetitle]{labeltitle}}
           {\printfield{label}}%
         \setunit{\printdelim{nonameyeardelim}}}
        {\printnames{labelname}%
         \setunit{\printdelim{nameyeardelim}}}%
       \iffieldundef{labelyear}
         {}
         {\printlabeldateextra}}
      {\printfield{shorthand}}}}

\renewbibmacro*{citeyear}{%
  \printtext[bibhyperref]{%
    \iffieldundef{shorthand}
      {\iffieldundef{labelyear}
         {\iffieldundef{label}
            {\printfield[citetitle]{labeltitle}}
            {\printfield{label}}}
         {\printlabeldateextra}}
      {\printfield{shorthand}}}}

\renewbibmacro*{textcite}{%
  \ifnameundef{labelname}
    {\iffieldundef{shorthand}
       {\usebibmacro{cite:label}%
        \setunit{%
          \global\booltrue{cbx:parens}%
          \printdelim{nonameyeardelim}\bibopenparen}%
        \ifnumequal{\value{citecount}}{1}
          {\usebibmacro{prenote}}
          {}%
        \usebibmacro{citeyear}}
       {\usebibmacro{cite:shorthand}}}
    {\printtext[bibhyperref]{\printnames{labelname}}%
     \setunit{%
       \global\booltrue{cbx:parens}%
       \printdelim{nameyeardelim}\bibopenparen}%
     \ifnumequal{\value{citecount}}{1}
       {\usebibmacro{prenote}}
       {}%
     \usebibmacro{citeyear}}}

\renewbibmacro*{in:}{}

\DeclareFieldFormat{pages}{#1}

\DeclareFieldFormat[article]{number}{\mkbibparens{#1}}

\renewbibmacro*{volume+number+eid}{%
  \printfield{volume}%
  \printfield{number}%
  \setunit{\bibeidpunct}%
  \printfield{eid}}

\AtEveryBibitem{
  \clearfield{doi}
  \clearfield{url}
}

\usepackage{bookmark}

\hypersetup{
  pdftitle={Multi-Armed Bandits with Arriving Arms: Sequential Screening, Dynamic Regret, and Sublinear Guarantees},
  pdfauthor={Deqi Zheng; Xiaoyang Xu; Yuhong Yang},
  pdfsubject={arXiv preprint},
  pdfkeywords={multi-armed bandits; arriving arms; dynamic regret; sequential screening; adaptive allocation}
}

\newcommand{\anon}{1}

\def\spacingset#1{%
  \renewcommand{\baselinestretch}{#1}\small\normalsize
}

\begin{document}

\def\spacingset#1{\renewcommand{\baselinestretch}%
{#1}\small\normalsize} \spacingset{1}

\if1\anon
{
  \title{\bf Multi-Armed Bandits with Arriving Arms: Sequential Screening, Dynamic Regret, and Sublinear Guarantees}
  \author{Deqi Zheng\thanks{
    The first two authors contributed equally to this work.}\\
    Qiuzhen College, Tsinghua University \\
    and \\
    Xiaoyang Xu$^{\ast}$ \\
    Qiuzhen College, Tsinghua University\\
    and \\
    Yuhong Yang\thanks{
    Corresponding author. Email: yyangsc@mail.tsinghua.edu.cn.}\hspace{.2cm}\\
    Yau Mathematical Sciences Center, Tsinghua University
    }
  \maketitle
} \fi

\if0\anon
{
  \bigskip
  \bigskip
  \bigskip
  \begin{center}
    {\LARGE\bf Multi-Armed Bandits with Arriving Arms: Sequential Screening, Dynamic Regret, and Sublinear Guarantees}
\end{center}
  \medskip
} \fi

\bigskip
\begin{abstract}
We study a stochastic multi-armed bandit problem in which the set of available arms expands over time. This setting arises in sequential experimentation when new actions or treatments become available during an ongoing study, making regret against a single best arm in hindsight inappropriate. We instead evaluate performance relative to the best arm currently available, leading to a dynamic-regret criterion for arriving-arm environments. To address the resulting challenges of arrival information discrepancy (AID) and a drifting benchmark (DB), we propose UCB for Arriving Arms (UCB-AA), an elimination-based procedure with an aiding preliminary screening step for newly arrived arms before full competition with incumbent arms. We show that UCB-AA attains regret bounds that depend explicitly on the arrival process, achieves sublinear dynamic regret under regularity conditions on gap evolution, and admits an online extension for unknown horizons. Simulation results show that UCB-AA reduces wasted pulls and maintains a smaller active arm set while preserving competitive regret performance.
\end{abstract}

\noindent%
\vfill

\newpage
\spacingset{1.15} 

\section{Introduction}

The classical multi-armed bandit framework assumes that the set of available actions is fixed in advance. In many modern applications, however, the action set expands over time. Precision oncology provides a representative example: as genomic profiling has advanced, the set of approved targeted therapies for non-small cell lung cancer has grown substantially, so a trial designed around a fixed collection of treatments may become outdated before it is completed. Similar issues arise in online advertising and other adaptive decision problems in which new options appear sequentially.

\begin{figure}[ht]
    \centering
    \includegraphics[width=0.9\textwidth]{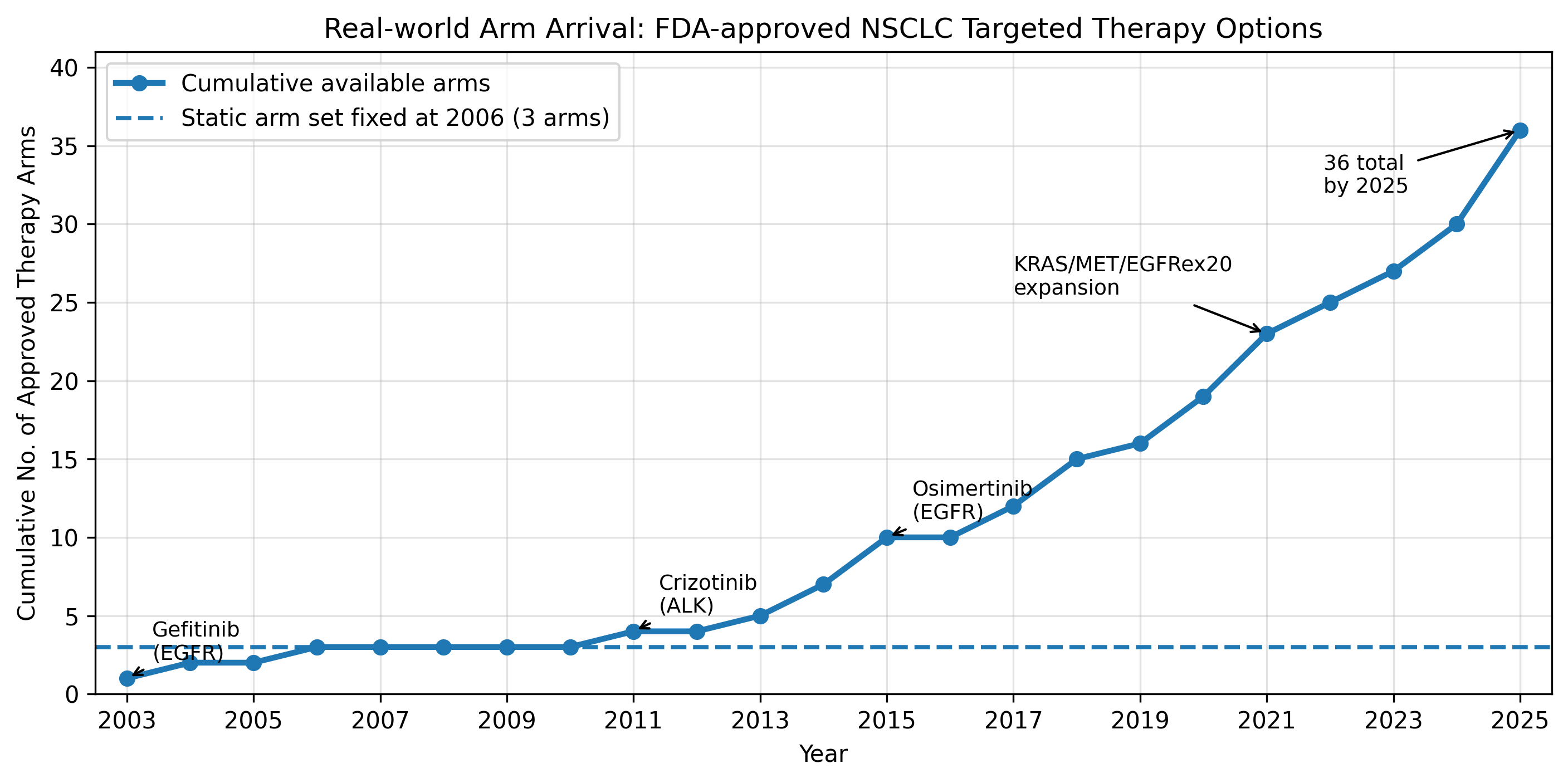}
    \caption{\textbf{The rapid expansion of the decision space in precision oncology.} FDA approval data for NSCLC (2003--2025). The solid line shows the cumulative number of targeted and immunotherapies (arms), from 1 in 2003 (Gefitinib) to 36 by 2025. A static algorithm fixed to the 2006 arm set would fail to incorporate most subsequently available treatments, illustrating the need for methods that accommodate arriving arms.}
    \label{fig:nsclc_growth}
\end{figure}

Motivated by such problems, we study the \textbf{multi-armed bandit problem with arriving arms} (MAB-AA). Let \(A_t\) denote the set of arms available at time \(t\), and suppose that \(A_t \subseteq A_{t+1}\), so that new arms arrive over time and remain available thereafter. In this setting, performance is more naturally evaluated against the best arm currently available than against the single best arm in hindsight, since the latter may not have been available earlier.

Relative to the classical stochastic bandit setting, MAB-AA introduces two main difficulties. The first is \emph{arrival information discrepancy (AID)}: newly arrived arms must be explored, which repeatedly diverts samples away from well-understood incumbent arms. The second is a \emph{drifting benchmark (DB)}: as better arms enter the system, both the identity of the optimal arm and the regret benchmark may change over time.

The arriving-arm setting is related to, but distinct from, several existing bandit formulations. Unlike infinite-armed bandits \parencite{10.1214/aos/1069362389}, where arms are actively sampled from a reservoir, new arms here arrive exogenously and remain available thereafter. Unlike mortal bandits \parencite{NIPS2008_788d9869}, the defining feature is not arm death, and unlike sleeping bandits \parencite{10.1007/s10994-010-5178-7}, the available set evolves monotonically over time. Our setting is also related to bandits with moving comparators: in nonstationary and switching bandits, the best arm may change because reward distributions vary over time \parencite{Besbes2014,Auer2019ADSWITCH}, whereas in MAB-AA each arm has a stationary reward law after arrival and benchmark changes are driven only by the exogenous arrival of new arms. 

Most closely related, \textcite{ghalme2021ballooning} studies ballooning bandits with regret measured against the best currently available arm, but focuses on feasibility of sublinear regret under restricted arrival patterns. Relatedly, \textcite{qi2025graphfeedback} considers an expanding-arm extension in graph-feedback bandits with similar arms. In contrast, our focus is the unstructured arriving-arm case, which is more suitable in applications where new options are introduced over time, previously introduced options remain available, and reliable similarity or contextual structure across options is unavailable. Examples include online product or content recommendation with continuously added items, adaptive A/B testing with newly launched treatments or designs, and sequential decision problems in which candidate actions are generated by an external pipeline rather than chosen by the learner.

From a statistical perspective, multi-armed bandit problems have long been studied as models for sequential allocation and adaptive experimentation, dating back to \textcite{Robbins1952} and the asymptotically efficient allocation theory of \textcite{LAI19854}; see also \textcite{Lai1987} for connections to adaptive treatment allocation. This viewpoint has motivated a broad statistical literature, including Bayesian and index-based approaches to sequential allocation \parencite{Gittins1979,BerryFristedt1985}, response-adaptive randomization in clinical trials \parencite{Zelen1969,WeiDurham1978,VillarBowdenWason2015,WilliamsonVillar2020,AzizKaufmannRiviere2021}, covariate-dependent and nonparametric bandit methods \parencite{Woodroofe1979,YangZhu2002,RigolletZeevi2010,PerchetRigollet2013}, and recent high-dimensional, semiparametric, and transfer-learning formulations \parencite{QianYang2016,QianIngLiu2023,CaiCaiLi2024}. These works share with ours the broad goal of learning from accumulated evidence to reduce allocation to inferior alternatives, but they focus on a fixed set of treatments or actions, and thus are not applicable to the arising arm setting that we study.

Methodologically, the two challenges call for different but complementary design principles. To address AID, our approach combines elimination steps \parencite{EvenDar2006SE,Auer2010UCBRI,PerchetRigollet2013,QianYang2016} with a pre-elimination mechanism tailored to the expanding action set. The key idea is to prevent the main competition stage from being repeatedly disrupted by every newly arriving arm. Instead, late arrivals are first screened against incumbent candidates, so that clearly inferior arms can be removed before receiving the full exploration budget assigned to serious contenders. To handle DB, the algorithm repeatedly updates the relevant comparison set as new arms arrive, and the analysis evaluates regret against the best arm currently available rather than against a fixed hindsight oracle. 

Based on these ideas, we propose \textbf{UCB for Arriving Arms (UCB-AA)}. For UCB-AA, we derive regret upper bounds that depend explicitly on the arrival process, establish sublinear dynamic regret under suitable conditions on gap evolution, study representative lower-bound regimes, and develop an online extension for unknown horizons.

Our contributions are threefold. First, we formulate the arriving-arm stochastic bandit problem using a dynamic-regret benchmark relative to the best currently available arm. Second, we propose a screening-based elimination procedure that is tailored to the information imbalance between incumbent and newly arrived arms. Third, we provide regret guarantees, lower-bound insights in representative regimes, and an online extension, together with simulations showing that UCB-AA reduces wasted pulls and maintains a compact active arm set.

The remainder of the paper is organized as follows. Section~\ref{sec:problem} formulates the problem and introduces the regret criterion. Section~\ref{sec:algorithm} presents UCB-AA, its theoretical guarantees, and an online extension for unknown horizons. Section~\ref{sec:lowerbound} develops lower bounds in representative regimes. Section~\ref{sec: simulation} reports numerical results. Section~\ref{sec:Discussion} concludes.

\section{Multi-Armed Bandit Problem with Arriving Arms }
\label{sec:problem}

\subsection{Problem Formulation}

We consider a sequential decision process over time steps $t = 1, 2, \dots$. Let $\mathcal{A}$ denote the (possibly infinite) set of all arms. At each time step $t$, only a subset $\mathcal{A}_t \subset \mathcal{A}$ is available, and we assume that the available set is non-decreasing:
$\mathcal{A}_t \subseteq \mathcal{A}_{t+1}.$
For any arm $i \in \mathcal{A}$, define its arrival time as
$\tau_i := \inf\{t\ge 1: i \in \mathcal{A}_t\}$.
Each arm $i$ is associated with an unknown reward distribution on $[0,1]$ with mean $r_i$. When arm $i$ is pulled at time $t$, a reward $X_{t,i}$ is generated; for each fixed $i$, we assume $\{X_{t,i}\}_{t\ge 1}$ are i.i.d. with mean $r_i$, and rewards are independent across arms. The arrival process is treated as exogenous to the learner.

A policy is a sequence $\pi = (\pi_t)_{t\ge 1}$, where each decision rule $\pi_t$ maps the past history to an available arm $I_t := \pi_t(H_{t-1}) \in \mathcal{A}_t,$ where $H_{t-1} := \big(\mathcal{A}_s, I_s, X_{s,I_s}\big)_{s=1}^{t-1}$
denote the observed history up to time $t-1$. Then the learner observes only $X_{t,I_t}$. Because the available set changes over time, the best arm may also change. We define the best mean reward at time $t$ as
$r^*(t) := \max_{i \in \mathcal{A}_t} r_i.$

For any (possibly random) stopping time $\tau$, define the cumulative regret by the (expected) regret against the best available arm at each time:
\[
R_{\tau}(\pi):=\sum_{t=1}^{\tau}\bigl(r^*(t)-r_{I_t}\bigr).
\]
For a deterministic horizon $T$, this reduces to $R_T(\pi)$. For the round-based procedure analyzed below,
the relevant stopping time is $T_0(\pi)$, the total number of pulls used to complete $M$
rounds. 
This benchmark differs from the usual regret against the single best arm in hindsight. In the arriving-arm setting, the terminal best arm in \(\mathcal{A}_T\) may not have been available in earlier periods, so comparing the learner with that arm throughout the entire horizon may overstate the achievable performance. The criterion above instead compares the learner at each time \(t\) with the best arm that was actually available at that time. When \(\mathcal{A}_t\) is constant over time, this formulation reduces to the classical stochastic multi-armed bandit problem.

\subsection{Main Statistical Challenges}
Relative to the classical fixed-arm stochastic bandit model, the arriving-arm setting introduces two related sources of difficulty.

\textbf{Arrival Information Discrepancy (AID):}
A first difficulty is arrival information discrepancy. With a finite sampling budget up to time $T$, the learner must allocate samples both to incumbent arms and to newly arriving arms whose rewards are still highly uncertain. Under continual arrivals, this repeated re-exploration can prevent the learner from concentrating pulls on the best currently available arm. Consequently, without additional regularity conditions on the arrival process, sublinear regret may be unattainable.

\textbf{Drifting Benchmark (DB):}
In contrast to the classical setting where the best mean reward is fixed, the benchmark in MAB-AA is time-dependent: $r^*(t)=\max_{i\in \mathcal{A}_t} r_i$. Since $\mathcal{A}_t$ expands over time, both the identity of the best arm and the value $r^*(t)$ may change when new arms arrive. Consequently, for a fixed arm $i$, the suboptimal gap $r^*(t)-r_i$ is generally time-varying (and non-decreasing in $t$). This drifting benchmark complicates both algorithm design and analysis, because the learner must continuously compare well-sampled incumbent arms with newly arriving arms whose estimates are still highly uncertain.

\section{UCB for Arriving Arms (UCB-AA) Algorithm}
\label{sec:algorithm}

\subsection{Algorithm Description}

UCB-AA is an elimination-based procedure organized by rounds. Its main idea is simple:
newly arriving arms are not allowed to compete immediately with well-sampled incumbent
arms. Instead, they are first buffered and screened at lower cost. Only those that survive
this screening step enter full competition in the next stage. This design aims to reduce
wasted exploration on clearly suboptimal late arrivals while preserving the ability to detect
genuinely competitive new arms.

Fix a finite number of rounds \(M\). For each round \(m=0,1,\dots,M\), let \(A_m\) denote
the set of arms arriving during round \(m-1\), with \(A_0\) the initial arm set, and define
\(
A(m):=\bigcup_{j=0}^m A_j .
\)
Let \(C_m\) denote the set of arms surviving round \(m\). Then round \(m\) starts from the
surviving incumbent arms \(C_{m-1}\) together with the newly admitted batch \(A_m\).
We write \(B_m\) for the active set in that round.

We assume that the cumulative number of arrived arms is bounded by a known nondecreasing
function \(K(m)\):
\[
|A(m)| \le K(m), \qquad |A_{m+1}| \le K(m+1)-K(m).
\]
For simplicity, assume that the best arm is unique in each round. Let
\(
r_{*(m)}:=\max_{i\in B_m} r_i,
\)
and let \(*(m)\in B_m\) denote the corresponding optimal arm. For each arm \(i\), define
its arrival round $t_i:=\min\{m:i\in A_m\}$, and its round-\(m\) suboptimality gap
\[
\Delta_i(m):=
\begin{cases}
r_{*(m)}-r_i, & m\ge t_i,\\
0, & m<t_i.
\end{cases}
\]

In round $m$, UCB-AA proceeds in two stages. As shown in Algorithm~\ref{alg:ucb-aa}, the earlier phases $p=0,\ldots,m-2$ are used for pre-elimination of newly arrived arms, while the last two phases $p=m-1,m$ are used for elimination on the merged active set.

\begin{algorithm}[H]
\caption{\textbf{UCB for Arriving Arms (UCB-AA)}}
\label{alg:ucb-aa}
\small
\begin{algorithmic}[1]
\Require number of rounds $M$, parameter $N>0$.
\State \textbf{Initialization:} $\tilde{\Delta}_0 \coloneqq 1$, $B_{-1} \coloneqq \varnothing$, $C_{-1} \coloneqq \varnothing$, $T_0 \gets 0$.

\Procedure{Elim}{$S,\;R,\;p$}
    \State $n_p \coloneqq \left\lceil \frac{2\log(N\tilde{\Delta}_p^2)}{\tilde{\Delta}_p^2} \right\rceil$
    \State Pull each $j\in S$ to $n_p$ total pulls; update $T_0$ after each pull
    \State Let $b_p\coloneqq\sqrt{\frac{\log(N\tilde{\Delta}_p^2)}{2n_p}}$; set $\mathrm{LCB}_j=\hat r_j(n_p)-b_p$, $\mathrm{UCB}_j=\hat r_j(n_p)+b_p$
    \State Remove from $S$ any $j$ with $\mathrm{UCB}_j \le \max\limits_{i\in R}\mathrm{LCB}_i$
    \State \Return $S$
\EndProcedure

\For{$m=0,1,\dots,M$}
    \State Receive $A_m$ and set $B_m^0 \coloneqq C_{m-1}\cup A_m$

    \If{$|A_m|\neq 0$ and $m\ge 2$}
        \State $A_m^0 \coloneqq A_m$
        \For{$p=0,1,\dots,m-2$}
            \State $A_m^{p+1} \gets$ \Call{Elim}{$A_m^p,\;B_m^p,\;p$}; $B_m^{p+1} \coloneqq C_{m-1}\cup A_m^{p+1}$
        \EndFor
    \EndIf

    \If{$m=0$}
        \State $B_m^{1} \gets$ \Call{Elim}{$B_m^0,\;B_m^0,\;0$}
    \Else
        \For{$p=m-1,m$}
            \State $B_m^{p+1} \gets$ \Call{Elim}{$B_m^p,\;B_m^p,\;p$}
        \EndFor
    \EndIf
    \State $C_m \coloneqq B_m^{m+1},\; \tilde{\Delta}_{m+1}=\frac{1}{2}\tilde{\Delta}_{m}$
\EndFor
\end{algorithmic}
\end{algorithm}

\subsubsection{Stage I: Pre-elimination of newly arrived arms}

At the beginning of round \(m\), the newly admitted batch \(A_m\) is screened
against the surviving incumbent set \(C_{m-1}\). During the preliminary phases
\(p=0,\ldots,m-2\), only arms in \(A_m\) are subject to removal: each surviving
new arm is sampled to the phase-specific target \(n_p\), and is discarded if its
upper confidence bound falls below the largest lower confidence bound of the
reference set.
This stage exploits the information asymmetry between incumbents and arrivals.
Incumbent arms have typically accumulated more samples, whereas newly arrived
arms are still highly uncertain. Screening arrivals before full competition
therefore prevents clearly inferior late arrivals from consuming the exploration
budget assigned to serious contenders.

\subsubsection{Stage II: Formal elimination on the active set}

After pre-elimination, the surviving new arms are merged with the incumbent arms to form
the active set for round \(m\). UCB-AA then applies elimination to the full active set.
At this stage, both incumbent and newly arrived arms may be removed. The surviving set is
denoted by \(C_m\) and is carried into the next round. Thus, unlike a naive elimination rule, UCB-AA does not allow newly arrived arms to enter the main elimination stage immediately upon arrival. They must first pass the preliminary screening step in phases $p=0,\ldots,m-2$. Once merged into the active set, however, all surviving arms are handled in the same way.

\subsection{Regret Decomposition}
\label{sec: Regret Decomposition}
Our analysis is indexed by the number of rounds \(M\), rather than by fixing the total
number of pulls in advance. We assume that the sampling budget is sufficient for UCB-AA
to complete all \(M\) rounds, so the total number of pulls used up to the end of round \(M\)
is a policy-dependent random variable. For a policy \(\pi\), let \(T_0(\pi)\) denote this total,
and write
\[
T:=\mathbb{E}[T_0(\pi)].
\]
Let \(N^\pi_{i,m}\) be the number of times arm \(i\) is
pulled during round \(m\). Then \(
T_0(\pi)=\sum_{m=0}^M \sum_{i\in B_m} N^\pi_{i,m},\)
where \(B_m\) is the active set at the beginning of round \(m\).

Because UCB-AA updates its active set only at round boundaries, it is convenient to first
analyze regret against the benchmark fixed at the beginning of each round. Define the
\emph{roundwise regret}
\[
R_M(\pi):=\sum_{m=0}^M \sum_{i\in B_m} N^\pi_{i,m}\Delta_i(m).
\]
This quantity charges each pull in round \(m\) against the best arm in the active set at the
start of that round.
Our target criterion, however, is the original dynamic regret
\[
R_{T_0}(\pi)=\sum_{t=1}^{T_0(\pi)}\bigl(r^*(t)-r_{I_t}\bigr),
\]
where \(r^*(t)=\max_{i\in A_t} r_i\) is the best mean reward among all arms available at
time \(t\). These two regret notions differ because a benchmark-improving arm may arrive
within a round but is not incorporated into the active set until the next round.

Accordingly, we decompose the dynamic regret as
\(
R_{T_0}(\pi)=R_M(\pi)+D_M(\pi),
\)
where
\[
D_M(\pi):=R_{T_0}(\pi)-R_M(\pi)
\]
is the \emph{delay regret}. The term \(R_M(\pi)\) captures the main statistical cost of
identifying competitive arms under the round structure, while \(D_M(\pi)\) captures the
additional loss caused by delayed incorporation of within-round benchmark improvements.
The remainder of this section controls these two terms separately: we first bound the
roundwise regret \(R_M(\pi)\), and then bound the delay regret \(D_M(\pi)\).

\subsection{Upper Bound for the Roundwise Regret}\label{sec: Upper Bound for the Roundwise Regret}

We now bound the principal term in the decomposition of Section 3.2.
The main question is how much regret is incurred before a suboptimal arm is removed under
the elimination schedule of UCB-AA.
To formalize this, define
\[
m_i:=\min\Bigl\{m\in\{t_i,t_i+1,\dots,M\}:\tilde\Delta_m<\tfrac12\Delta_i(m)\Bigr\},
\]
the first round in which arm \(i\) becomes statistically eliminable under the target
resolution sequence. Also define
\[
\tilde m_i:=\min\Bigl\{m\in\{0,1,\dots,M\}:\tilde\Delta_m<\tfrac12\Delta_i(m_i)\Bigr\},
\]
so that \(\tilde m_i\le m_i\).
We classify suboptimal arms according to the phase in which they are eliminated with high
probability:
\begin{itemize}
    \item \(G_1\) ($m_i=t_i$): arms eliminated when they first appear;
    \item \(G_2\) ($m_i>t_i,\;\tilde{m}_i=m_i$): arms are eliminated in the formal elimination phase \(p=m_i\) of round \(m_i\);
    \item \(G_3\) ($m_i>t_i,\;\tilde{m}_i<m_i$): arms are eliminated in the formal elimination phase \(p=m_i-1\) of round \(m_i\).
\end{itemize}

For a threshold \(\lambda \ge 2^{-M+1}\), define
\(
A'(\lambda):=\{i\in A(M):\Delta_i(M)>\lambda\}.
\)
Then \(A'(\lambda)\) admits the partition
\(
A'(\lambda)=G_1(\lambda)\cup G_2(\lambda)\cup G_3(\lambda).
\)
Moreover, define
\[
\bar m_i:=
\begin{cases}
\tilde m_i, & i\in G_1(\lambda)\cup G_2(\lambda),\\
m_i-1, & i\in G_3(\lambda).
\end{cases}
\]
Thus \(\bar m_i\) is the elimination scale at which arm \(i\) is removed with high probability.

\begin{theorem}[Upper bound for the roundwise regret]
\label{thm:ucb-aa-upper}
For any $\lambda\geq 2^{-M+1}$, the regret of UCB-AA up to round $M$ satisfies
\begin{align*}
\mathbb{E}R_M(\pi)
\leq
\sum_{\substack{i: \Delta_i(M) \geq \lambda}}
\left(
\tilde{S}_i
+
\frac{4\bigl(K(m_i)+5\bigr)\,R_i}{N\tilde{\Delta}_{\tilde{m}_i}^2}
\right)
+
T\max_{i:\Delta_i(M)\leq \lambda}\Delta_i(M),
\end{align*}
where $T = \mathbb{E}[T_0(\pi)]$ denotes the expected total number of pulls,
\[
R_i\coloneqq n_{t_i}\Delta_i(t_i)+\sum_{m=t_i+1}^M\bigl(n_m-n_{m-1}\bigr)\Delta_i(m),
\]
and
\[
\tilde{S}_i\coloneqq
\begin{cases}
    n_{\tilde{m}_i}\,\Delta_i(t_i), & i\in G_1(\lambda),\\
    n_{t_i}\Delta_i(t_i)+\sum_{m=t_i+1}^{\overline{m}_i}\bigl(n_m-n_{m-1}\bigr)\Delta_i(m), & i\in G_2(\lambda)\cup G_3(\lambda).
\end{cases}
\]
\end{theorem}

Theorem~\ref{thm:ucb-aa-upper} decomposes the roundwise regret into three statistically interpretable components. The leading term, \(\tilde S_i\), is the identification cost of certifying that a suboptimal arm should be removed at the resolution dictated by the elimination schedule. The second term quantifies the additional regret incurred on low-probability events where statistical fluctuations delay elimination or allow an incorrect survivor to persist. The final truncation term collects arms whose terminal gaps are below the target resolution \(\lambda\), and may therefore be treated as effectively indistinguishable at the scale of the analysis.

To obtain a simpler rate statement, we impose the following stability condition.

\begin{assumption}[Gap stability]
\label{ass:gap_stability}
For every suboptimal arm \(i \in A'(\lambda)\), there exists a universal constant \(\tilde{C}>0\), independent of \(\lambda\), such that
\[
\Delta_i(M) \leq \tilde{C}\,\Delta_i(m_i).
\]
\end{assumption}

Assumption~\ref{ass:gap_stability} does not restrict the arrival times of arms;
rather, it limits how much the benchmark can improve after an arm has already
become statistically distinguishable. Equivalently, the post-detection benchmark
increase cannot be larger than a constant multiple of the gap already visible at
\(m_i\).
This is a mild condition in several common regimes. It holds with
\(\tilde C=1\) if the best arm available at the terminal round has already
arrived by \(m_i\), or more generally if no later arrival improves the benchmark.
It also holds whenever later improvements are moderate relative to the current
detectable gap, for example if
\(
r_{*(M)}-r_{*(m_i)}\le c\,\Delta_i(m_i)
\)
for a constant \(c\), in which case \(\tilde C=1+c\). This covers environments
with bounded multiplicative benchmark growth after detection, as well as
settings in which only a finite number of post-detection benchmark-improving
arrivals occur and their cumulative improvement is comparable to the detection
scale. Another simple case is a separated-gap regime: if every arm in
\(A'(\lambda)\) has \(\Delta_i(m_i)\ge \delta_0>0\) for a fixed \(\delta_0\),
then bounded rewards imply \(\Delta_i(M)\le 1\le \delta_0^{-1}\Delta_i(m_i)\).
Thus the assumption mainly rules out extreme post-detection drift, where an arm
is already clearly inferior but many substantially better arms arrive later and
inflate its terminal gap by an unbounded factor.

\begin{corollary}
\label{cor:ucb_aa_simplified}
Suppose that Assumption~\ref{ass:gap_stability} holds for all $i\in A'(\lambda)=\{i\in A(M):\Delta_i(M)>\lambda\}$. Let $K\coloneqq K(M)$ and choose
\[
N = 10K\,n_M,
\qquad
\lambda \asymp \sqrt{\frac{K\log K+K}{T}}.
\]
Then the roundwise regret of UCB-AA satisfies
\[
\mathbb{E}R_M(\pi)\lesssim \sqrt{K\log K\cdot T}.
\]
\end{corollary}

Corollary~\ref{cor:ucb_aa_simplified} shows that UCB-AA achieves sublinear roundwise regret when the effective
number of arriving competitors remains moderate relative to the available sample size.
Here \(K=K(M)\) is the cumulative number of arms that may have appeared by round \(M\),
rather than the total size of the global arm universe. In particular, the bound is informative
in regimes where \(
K(M)=o\left(\frac{T}{\log T}\right),
\)
so that the AID effect created by continual arrivals remains controllable.

\begin{note}
The parameter \(N\) must be fixed before running UCB-AA. In the theoretical analysis,
we use the conservative calibration
\(
N=10K(M)n_M, 
\;\;
n_M=\left\lceil 
\frac{2\log(N\widetilde\Delta_M^2)}{\widetilde\Delta_M^2}
\right\rceil ,
\)
where \(K(M)\) is a prior upper bound on the cumulative number of arrived arms. Although
this implicit relation can be solved approximately, its role is mainly analytical: it ensures
uniform control of low-probability failure events over all arms, but may lead to overly large
pull targets \(n_m\), and hence excessive forced exploration in both the pre-elimination and
formal elimination stages. For implementation, we use a less conservative calibration based only on the monotonicity
of the target sequence \(\{n_m\}_{m=0}^M\). Since
\(
f(x)=\frac{\log(Nx)}{x}
\)
is decreasing on \([\widetilde\Delta_M^2,1]\) whenever
\(N\widetilde\Delta_M^2\ge e\), a natural lower-end choice is
\(
N_{\min}=\frac{e}{\widetilde\Delta_M^2}.
\)
This optimistic calibration is not used in the regret proof, but provides a practical baseline
with substantially smaller pull targets, and is adopted in the simulations.

\end{note}

\subsection{Control of the Delay Regret}

Delay regret arises because UCB-AA updates its active set only at round boundaries. If a
benchmark-improving arm arrives in the middle of round \(k\), then the true dynamic
benchmark improves immediately on the original timeline, whereas the roundwise benchmark
used in \(R_M(\pi)\) is updated only at the beginning of round \(k+1\). Thus \(D_M(\pi)\)
captures the additional loss caused by this deferred incorporation of within-round benchmark
improvements.

Fix a round \(k\). Suppose that within this round there are \(q_k\) benchmark-improving
arrival events, with increments
\[
\beta_{k,1},\dots,\beta_{k,q_k}\ge 0,
\qquad 
\sum_{j=1}^{q_k}\beta_{k,j}=: \delta_k,
\]
where
\(
\delta_k=r_{*(k+1)}-r_{*(k)}
\)
is the total benchmark increase associated with round \(k\). Let \(L_k\) be the total number
of pulls in round \(k\), and let
\(
1\le \tau_{k,1}<\cdots<\tau_{k,q_k}\le L_k
\)
denote the within-round locations of these improving arrivals. We define the exposure factor
\[
\rho_k:=
\frac{1}{L_k\delta_k}
\sum_{j=1}^{q_k}(L_k-\tau_{k,j}+1)\beta_{k,j},
\qquad \delta_k>0,
\]
so that \(0\le \rho_k\le 1\). The factor \(\rho_k\) measures how much of round \(k\) is exposed
to the benchmark increase: improvements that occur early contribute more to delay regret than
those that occur late.
For \(\ell\ge k+1\), define the cumulative benchmark improvement after round \(k\) by
\[
\Delta_{*(k)}(\ell):=r_{*(\ell)}-r_{*(k)}=\sum_{i=k}^{\ell-1}\delta_i.
\]
We also define
\(
m_{*(k)}=\min\Bigl\{\ell\ge k+1:\tilde\Delta_\ell<\tfrac12\Delta_{*(k)}(\ell)\Bigr\},
\)
the first round at which the cumulative improvement after round \(k\) becomes statistically
detectable under the elimination schedule.

\begin{note}
For analytical convenience, we take the analyzed horizon to end at a checkpoint immediately
after the last batch of arrivals has been admitted. Equivalently, one may append a terminal
bookkeeping round with no further sampling. This convention does not require the learner to
determine whether the final arrivals improve the benchmark; it only ensures that no arrivals
remain unincorporated at the endpoint. Under this convention, the terminal delay contribution
is null, so the endpoint condition involving \(\rho_M\delta_M\) becomes vacuous.
\end{note}

To bound \(D_M(\pi)\), we impose a condition that limits how aggressively benchmark
improvements can accumulate before they become statistically detectable. This condition is
used to close the upper bound and should be viewed as a sufficient condition, not a necessary
one.

\begin{assumption}[Cumulative jump control]\label{ass:delay_cum}
Assume that for every round $k$ satisfying $\rho_k\delta_k>\lambda$, one has $m_{*(k)}>k+1$,
and at least one of the following conditions holds:
\begin{align}
\Delta_{*(k)}\bigl(m_{*(k)}\bigr)
&\le
\tilde{\Delta}_k
\sqrt{
\frac{M-m_{*(k)}+1}{3\rho_k K(k)(M-k+1)}
},
\label{eq:delay-cond-1}
\\
\frac{\delta_k}{\Delta_{*(k)}(m_{*(k)}-1)}
&\le
\frac{2^{\,m_{*(k)}-k-2}}{\rho_k K(k)}.
\label{eq:delay-cond-2}
\end{align}
\end{assumption}
Assumption~\ref{ass:delay_cum} is most naturally satisfied when benchmark improvements remain moderate until
they become detectable at the current resolution scale. This includes, for example, settings in
which improvements occur gradually, settings in which large improvements are rare, or settings
in which most benchmark-improving arrivals occur late within a round and therefore have
small exposure factors \(\rho_k\).

\begin{proposition}[Delay regret under cumulative jump control]\label{prop:delay_cum}
Let
\(
\lambda\ge 2^{-M+1}.
\)
Under Assumption~\ref{ass:delay_cum},
\[
\mathbb{E}\bigl[D_M(\pi)\bigr]
\le
\sum_{i:\Delta_i(M)>\lambda}\tilde S_i+\lambda T.
\]
\end{proposition}

Proposition \ref{prop:delay_cum} shows that, under cumulative control of benchmark jumps, the delay regret is bounded by the same certification-type quantity that appears in the main regret analysis, up to the truncation term $\lambda T$. Hence the delayed admission of within-round arrivals does not create a new leading-order source of statistical difficulty. More broadly, the result clarifies how the dynamic benchmark enters the regret analysis: the regret of MAB-AA can be viewed as consisting of a principal identification term and an additional delay term induced by benchmark improvement. Under Assumption~\ref{ass:delay_cum}, the latter remains controlled, so benchmark movement acts as a structured perturbation of the elimination problem rather than as a fundamentally separate source of regret.

\subsection{General regret bound}

We now return to the original dynamic regret
\(
R_{T_0}(\pi)
=
\sum_{t=1}^{T_0(\pi)}\bigl(r^*(t)-r_{I_t}\bigr).
\)
By the decomposition in Section~\ref{sec: Regret Decomposition},
\[
R_{T_0}(\pi)=R_M(\pi)+D_M(\pi).
\]
Theorem~\ref{thm:ucb-aa-upper} controls the principal term \(R_M(\pi)\), and Proposition~\ref{prop:delay_cum} controls the delay term \(D_M(\pi)\). Combining the two results yields the following bound for the original regret criterion.

\begin{theorem}[General regret bound]
\label{thm:general bound}
Let 
\(
\lambda \ge 2^{-M+1}.
\)
Suppose that Assumptions \ref{ass:gap_stability} and \ref{ass:delay_cum} hold. Then the dynamic regret of UCB-AA satisfies
\[
\mathbb{E} R_{T_0}(\pi)
\le
\sum_{i:\Delta_i(M)\ge \lambda}
\left(
2\tilde S_i+
\frac{4\bigl(K(m_i)+5\bigr)R_i}{N\tilde\Delta_{\tilde m_i}^2}
\right)
+2\lambda T .
\]
In particular, under the conservative calibration of $N$ in Corollary~\ref{cor:ucb_aa_simplified},
\[
\mathbb{E}R_{T_0}(\pi)\lesssim \sqrt{K\log K\cdot T}.
\]
\end{theorem}

\subsection{Online UCB-AA}
\label{sec:online}

We briefly discuss the case in which the terminal round \(M\) is unknown. To remove the fixed-horizon requirement, we use a geometric-restart scheme and run UCB-AA over successive stages with geometrically increasing stage lengths. This yields an anytime version of the procedure, referred to as Online UCB-AA. The idea is standard: each stage is long enough to cover the possibility that the effective horizon is of that order, and restarting across stages incurs only a logarithmic overhead relative to the fixed-\(M\) analysis. We retain the algorithm here for completeness and defer the more detailed stage-wise analysis to the Supplementary Material.

\begin{algorithm}[htbp]
\caption{Online UCB-AA}
\label{alg:online-ucb-aa}
\begin{algorithmic}[1]
\Require Base algorithm UCB-AA
\State Initialize stage index $l \leftarrow 1$
\While{interaction continues}
    \State $M_l\leftarrow2^l$, $\tilde{M}_l\leftarrow\sum_{i=1}^l M_i$, $\tilde{\Delta}_{M_l}\leftarrow2^{-M_l}$
    \State $n_{M_l}(N_l)\coloneqq\left\lceil\frac{2\log(N_l\tilde{\Delta}_{M_l}^2)}{\tilde{\Delta}_{M_l}^2}\right\rceil$
    \State Compute $N_l$ from $N_l=10K(\tilde{M}_l)n_{M_l}(N_l)$ (approx.)
    \State Run UCB-AA$(M_l,N_l)$ for $M_l$ rounds
    \State $l \leftarrow l+1$
\EndWhile
\end{algorithmic}
\end{algorithm}

At each restart, Online UCB-AA reinitializes the internal state of the base
procedure for the new stage, while all pulls across stages are still counted in
the cumulative regret.
The next corollary summarizes the resulting regret guarantee in the unknown-horizon setting.

\begin{corollary}
\label{cor:online_gap_independent}
Suppose that Assumption~\ref{ass:gap_stability} holds.
The expected roundwise regret of Online UCB-AA satisfies
\[
\mathbb{E}R_M(\pi) \lesssim \sqrt{K(M)\,T\,\log T}.
\]
If a stagewise version of Assumption~\ref{ass:delay_cum} holds uniformly across the restart stages,
then
\[
\mathbb{E}R_{T_0}(\pi) \lesssim \sqrt{K(M)\,T\,\log T}.
\]
\end{corollary}

Corollary~\ref{cor:online_gap_independent} shows that the online extension preserves the same square-root-type behavior as the fixed-\(M\) procedure, at the cost of only a logarithmic overhead due to geometric restarts. Thus, lack of prior knowledge of the terminal round does not alter the qualitative statistical difficulty of the arriving-arm problem. A more detailed stage-wise bound is given in the Supplementary Material.

\section{Lower Bounds in Representative Regimes}
\label{sec:lowerbound}

As mentioned earlier, without structural restrictions on the arrival process, sublinear regret may be unattainable
\parencite{ghalme2021ballooning}. A fully general lower bound for the unrestricted arriving-arm model is therefore not our goal here. Instead, we focus on three representative regimes that isolate distinct hardness mechanisms in MAB-AA: static identification, repeated distractor certification, and near-tied benchmark updates under a drifting comparator.

The statistical hardness of MAB-AA is therefore not monolithic. In the static limit, the problem reduces to the classical identification barrier of stochastic bandits. When a globally optimal incumbent is present from the outset and all late arrivals are inferior, the dominant cost is repeated certification of distractors. When the benchmark improves through a sequence of near-tied arrivals, the difficulty becomes a succession of local tests under a drifting comparator. This section isolates these three mechanisms and relates them to the upper-bound structure in Section~\ref{sec:algorithm}.

To avoid repeating regime-specific growth conditions, we use the following scale-sensitive extension of the usual notion of uniform goodness.

\begin{definition}[Uniform goodness at scale $h(T)$]
\label{def:uniformly-good-scale}
Let $\mathcal V_T$ be a family of bandit instances indexed by $T$, and let $h(T)\to\infty$.
A sequence of policies $\{\pi^{(T)}\}_{T\ge1}$ is \emph{uniformly good on $\{\mathcal V_T\}$ at
scale $h(T)$} if, for every $a>0$,
\[
\sup_{\nu\in\mathcal V_T}\mathbb E_{\nu}[R_T(\pi^{(T)};\nu)]
=
o\bigl(h(T)^a\bigr)
\qquad\text{as }T\to\infty.
\]
\end{definition}

 When $h(T)=T$, it reduces to the classical requirement considered in \textcite{LAI19854,BurnetasKatehakis1996}. For related information-theoretic change-of-measure techniques in best-arm
identification, see also \textcite{KaufmannCappeGarivier2016}.

\subsection{Static Reduction}

We first consider the degenerate no-arrival case. Here $h(T)=T$, so
Definition~\ref{def:uniformly-good-scale} reduces to the usual notion of uniform goodness.
The purpose of this subsection is mainly calibrational: when no new arms enter, MAB-AA
reduces to the classical stochastic bandit problem, so the arriving-arm framework should
recover the familiar logarithmic identification barrier. Accordingly, we restate a standard
lower-bound implication from the classical bandit literature
\parencite{LAI19854,BurnetasKatehakis1996} in a form adapted to our notation. This result
is included for comparison with our arriving-arm analysis, rather than as a new contribution.
To avoid unnecessary generality, we use a local formulation under a quadratic KL condition,
which is sufficient for our comparison with the upper bounds in
Section~\ref{sec:algorithm}.

Consider a static bandit family with rewards supported on $[0,1]$. Let $\nu^{(0)}$ be a baseline
instance in which arm $0$ is the unique optimal arm with mean $r^\star\in(0,1)$. For
$i=1,\dots,K$, let arm $i$ have reward law $P_i$ and mean
\(
r_i=r^\star-\Delta_i\) where \(
\Delta_i>0.
\)

\begin{theorem}[Static lower bound]
\label{thm:static-lb-local-quadratic}
Assume that for each $i=1,\dots,K$ there exists a distribution $Q_i\in\mathcal P([0,1])$ such
that
\(
\eta_i:=\mathbb E_{Q_i}[X]-r^\star>0\) and
\(
D(P_i\|Q_i)\le C_i\Delta_i^2
\)
for some finite constant $C_i>0$. Let $\nu^{(i)}$ denote the instance obtained from $\nu^{(0)}$
by replacing only the reward law of arm $i$ from $P_i$ to $Q_i$. If a policy $\pi$ is uniformly
good on the local family $\mathcal V:=\{\nu^{(0)},\nu^{(1)},\dots,\nu^{(K)}\}$, then
\[
\mathbb E_{\nu^{(0)}}[R_T(\pi;\nu^{(0)})]
\ge
c\sum_{i=1}^K \frac{\log T}{C_i\Delta_i}
\]
for all sufficiently large $T$, where $c>0$ is an absolute constant. In particular, if $\sup_{1\le i\le K} C_i\le C_\star<\infty$, then
\[
\mathbb E_{\nu^{(0)}}[R_T(\pi;\nu^{(0)})]
\ge
c\,C_\star^{-1}\sum_{i=1}^K \frac{\log T}{\Delta_i}
\]
for all sufficiently large $T$.
\end{theorem}

\begin{example}[Static specialization]
\label{exp:static}
In the absence of arrivals, UCB-AA specializes to an elimination-style stochastic bandit
procedure and achieves
\(
\tilde O\left(\sum_i \frac{\log(T\Delta_i^2)}{\Delta_i}\right)
\)
instance-dependent regret. Together with
Theorem~\ref{thm:static-lb-local-quadratic}, this shows that the arriving-arm reduction does
not sacrifice statistical efficiency in the static limit, up to logarithmic refinements.
\end{example}

\subsection{Sequential Distractors}

We next consider a regime in which a globally optimal incumbent is present from the outset,
while every newly arriving arm is suboptimal under the baseline instance. The difficulty is
therefore repeated certification: each arrival creates a local alternative under which that arm
would become optimal, so the learner must accumulate logarithmic evidence before discarding it.

For each horizon $T$, let $\nu^{(0)}$ be a baseline instance in which arm $0$ is the unique
optimal arm, available from time $1$, with mean $r^\star\in(0,1)$. Let $\mathcal I_T$ be a
collection of arriving distractor arms. For each $i\in\mathcal I_T$, suppose arm $i$ arrives at
time $\tau_i\le T$, has reward law $P_i$, and mean
\(
r_i=r^\star-\Delta_i
\) where \(
\Delta_i>0.
\)
Write
\[
H_i(T):=T-\tau_i+1,
\qquad
H_{\min}(T):=\min_{i\in\mathcal I_T}H_i(T),
\]
and assume $H_{\min}(T)\to\infty$. In this regime we evaluate uniform goodness at the effective
scale $h(T)=H_{\min}(T)$.

\begin{theorem}[Distractor lower bound under local alternatives]
\label{thm:distractor-lb-local}
Suppose that for each $i\in\mathcal I_T$ there exists a law $Q_i\in\mathcal P([0,1])$ such that
\(
\eta_i:=\mathbb E_{Q_i}[X]-r^\star>0 \) and
\(
d_i:=D(P_i\|Q_i)<\infty.
\)
Let $\nu^{(i)}$ denote the instance obtained from $\nu^{(0)}$ by replacing the reward law
of arm $i$ from $P_i$ to $Q_i$, and set
\(
\mathcal V_T:=\{\nu^{(0)}\}\cup\{\nu^{(i)}:i\in\mathcal I_T\}.
\)
Assume that, for all sufficiently large $T$,
\[
\inf_{i\in\mathcal I_T}\Delta_i\ge \underline\Delta>0,
\qquad
\inf_{i\in\mathcal I_T}\eta_i\ge \underline\eta>0,
\qquad
\sup_{i\in\mathcal I_T}d_i\le \overline d<\infty.
\]
Then any sequence $\{\pi^{(T)}\}_{T\ge1}$ uniformly good on $\mathcal V_T$ satisfies
\[
\mathbb E_{\nu^{(0)}}[R_T(\pi^{(T)};\nu^{(0)})]
\ge
c\sum_{i\in\mathcal I_T}\frac{\Delta_i}{d_i}\log H_i(T)
\ge
c\,\underline\Delta\,\overline d^{-1}\,|\mathcal I_T|\,\log H_{\min}(T)
\]
for all sufficiently large $T$, where $c>0$ is an absolute constant.
\end{theorem}

\begin{corollary}[Two canonical distractor regimes]
\label{cor:distractor-lb-regimes}
Under the assumptions of Theorem~\ref{thm:distractor-lb-local}, suppose in addition that one
of the following holds:

\smallskip
\noindent
\textup{(i) Equal-distance arrival: } each arm in $\mathcal I_T=\{1,\dots,K_T\}$ has the same arrival interval $H:=\lfloor T/K_T\rfloor$ which means $\tau_i=(i-1)H+1$,and $H$ tends to infinity.

\smallskip
\noindent
\textup{(ii) Early arrival: } there exists a fixed $\alpha\in(0,1)$ such that
\(
\tau_i\le \alpha T\)
for every \(i\in\mathcal I_T.
\)

Then
\[
\mathbb E_{\nu^{(0)}}[R_T(\pi^{(T)};\nu^{(0)})]
\ge
c\sum_{i\in\mathcal I_T}\frac{\Delta_i}{d_i}\log H_{\min}(T)
\]
for all sufficiently large $T$. In particular, if
\[
\sup_{i\in\mathcal I_T} d_i\le \overline d<\infty,
\qquad
\inf_{i\in\mathcal I_T}\Delta_i\ge \underline\Delta>0,
\qquad
|\mathcal I_T|\asymp K_T,
\]
then
\[
\mathbb E_{\nu^{(0)}}[R_T(\pi^{(T)};\nu^{(0)})]
=
\Omega\bigl(K_T\log H_{\min}(T)\bigr).
\]
\end{corollary}

The lower bound reflects a repeated certification cost: although the arriving arms are inferior in
the baseline instance, a uniformly good policy cannot ignore them, because each arrival induces
a local alternative under which that arm is optimal.

\begin{example}[Fixed-gap distractors]
\label{exp:distractors}
Consider an initial set of arms $A_0=\{*,0\}$, where arm $*$ is globally optimal with mean
$\mu_*=\tfrac34$ and arm $0$ has mean $\mu_0=\tfrac14$. For each round $m=1,\dots,M$, a
single new arm $i=m$ arrives with
\(
\mu_i=\frac12-2^{-i-2}.
\)
Let $K(m)=m+2$ denote an upper bound on the number of arms available up to round $m$.
All new arrivals are distractors: their means converge to $1/2$, while the optimal arm remains
fixed at $3/4$, so the gaps stay bounded away from zero. Applying
Theorem~\ref{thm:ucb-aa-upper} with $N=10K(M)n_M$ yields
\(
\mathbb E R_T(\pi_{\mathrm{UCB\text{-}AA}})
=
\mathbb E R_M(\pi_{\mathrm{UCB\text{-}AA}})
=
\widetilde O\bigl(K(M)\log T\bigr).
\)
\end{example}

Combined with Theorem~\ref{thm:distractor-lb-local}, this gives the comparison
\[
\widetilde O\bigl(K_T\log T\bigr)
\qquad\text{versus}\qquad
\Omega\bigl(K_T\log(T/K_T)\bigr),
\]
so in the canonical sparse-arrival regime $K_T=o(T)$ the rates agree up to logarithmic factors.
A complementary failure-mode construction showing that the absence of pre-elimination can lead
to linear regret is deferred to the Supplementary Material.

\subsection{Progressive Near-tied Arrivals}

We finally consider a regime in which the benchmark improves through a sequence of near-tied
arrivals. Unlike the distractor setting, the main difficulty is no longer certifying clearly inferior
arms, but repeatedly deciding whether a new challenger lies slightly below or slightly above the
current reference level. Because this mechanism is intrinsically local and horizon dependent, the
natural lower bound here is a finite-horizon minimax statement.

Fix a horizon $T$ and integers $K_T\ge1$, $H_1,\dots,H_{K_T}\in\mathbb N$ such that
\(
\sum_{m=1}^{K_T} H_m=T.
\)
Define checkpoints and blocks by
\[
s_0:=0,
\qquad
s_m:=\sum_{\ell=1}^m H_\ell,
\qquad
\mathcal B_m:=\{s_{m-1}+1,\dots,s_m\},
\qquad
m=1,\dots,K_T.
\]
Let $\gamma\in(0,1/2]$, and choose
\(
\theta_1,\dots,\theta_{K_T}\in(0,1),
\;
\varepsilon_1,\dots,\varepsilon_{K_T}>0
\) 
such that
\begin{align}
\theta_m-\varepsilon_m,\ \theta_m,\ \theta_m+\varepsilon_m
&\in[\gamma,1-\gamma],
\qquad m=1,\dots,K_T,
\label{eq:prog-lb-interior}
\\
\theta_{m+1}
&\ge \theta_m+\varepsilon_m,
\qquad m=1,\dots,K_T-1.
\label{eq:prog-lb-monotone}
\end{align}
For each sign vector
\(
\sigma=(\sigma_1,\dots,\sigma_{K_T})\in\{-1,+1\}^{K_T},
\)
define an arriving-arm instance $\nu_\sigma$ in which, at time $s_{m-1}+1$, two Bernoulli
arms $c_m,d_m$ arrive and remain available thereafter, with means
\(
\mu(c_m)=\theta_m\) and \(
\mu(d_m)=\theta_m+\sigma_m\varepsilon_m.
\)
Let
\(
\mathcal V_T^{\mathrm{prog}}
:=
\{\nu_\sigma:\sigma\in\{-1,+1\}^{K_T}\}.
\)

\begin{theorem}[Deterministic-checkpoint minimax lower bound]
\label{thm:prog-lb-general}
Under \eqref{eq:prog-lb-interior}--\eqref{eq:prog-lb-monotone}, every non-anticipating policy
$\pi$ satisfies
\begin{align}
\sup_{\nu\in\mathcal V_T^{\mathrm{prog}}}
\mathbb E_\nu[R_T(\pi;\nu)]
\ge
\frac12
\sum_{m=1}^{K_T}
H_m\varepsilon_m
\left(
1-
\sqrt{
\frac{2H_m\varepsilon_m^2}{\gamma(1-\gamma)}
}
\right)_+ .
    \label{eq:prog-lb-kl-simple}
\end{align}
In particular, if there exists a constant $\kappa\in(0,1/2)$ such that
\[
H_m\varepsilon_m^2
\le
\kappa\,\gamma(1-\gamma),
\qquad
m=1,\dots,K_T,
\]
\begin{align}
    \text{then}\quad \sup_{\nu\in\mathcal V_T^{\mathrm{prog}}}
\mathbb E_\nu[R_T(\pi;\nu)]
\ge
c_\kappa
\sum_{m=1}^{K_T} H_m\varepsilon_m,
\qquad
c_\kappa:=\frac12(1-\sqrt{2\kappa})>0.
\label{eq:prog-lb-clean}
\end{align}
The same lower bounds remain valid after taking $\inf_\pi$ on the left-hand side.
\end{theorem}

Each block creates a fresh one-sided local test: the learner must determine whether the new
challenger lies slightly below or slightly above the current benchmark. When
$H_m\varepsilon_m^2$ is of constant order, the resulting regret contribution is of order
$H_m\varepsilon_m$, and these local-testing costs add across blocks.

\begin{corollary}[$\sqrt{K_TT}$ lower bound under near-homogeneous local tests]
\label{cor:prog-lb-sqrtKT}
Under the assumptions of Theorem~\ref{thm:prog-lb-general}, assume in addition that there
exist constants $c_H,C_H,c_\varepsilon,C_\varepsilon>0$ and $\kappa\in(0,1/2)$ such that, for
every $m=1,\dots,K_T$,
\[
c_H\,\frac{T}{K_T}
\le
H_m
\le
C_H\,\frac{T}{K_T},
\qquad
c_\varepsilon\sqrt{\frac{K_T}{T}}
\le
\varepsilon_m
\le
C_\varepsilon\sqrt{\frac{K_T}{T}},
\]
and
\(
C_HC_\varepsilon^2\le \kappa\,\gamma(1-\gamma).
\)
Then
\[
\inf_\pi\sup_{\nu\in\mathcal V_T^{\mathrm{prog}}}
\mathbb E_\nu[R_T(\pi;\nu)]
\ge
c\,\sqrt{K_TT}
\]
for all sufficiently large $T$, where $c>0$ depends only on $c_H,c_\varepsilon,$ and $\kappa$.
\end{corollary}

Corollary~\ref{cor:prog-lb-sqrtKT} identifies the canonical square-root minimax scale in a
near-homogeneous local-testing regime. Our general upper bound in
Section~\ref{sec:algorithm} matches this dependence up to logarithmic factors. The example
below is not a literal specialization of the deterministic-checkpoint lower bound, because
UCB-AA operates with endogenous random checkpoints, but it exhibits the same qualitative
mechanism.

\begin{example}[Monotone near-tied ladder]
\label{exp:monotone-near-tied-ladder}
Fix $M\ge 5$ and define $\xi_m := \frac12 - 2^{-m-2}$ for $m=0,1,\dots,M$. At round $0$,
there is a single arm $a_0$, so that $A_0=\{a_0\}$ and $\mu(a_0)=\xi_0$. For each round
$m=1,\dots,M$, two new Bernoulli arms arrive, namely $A_m=\{c_m,d_m\}$, with means
$\mu(c_m)=\xi_{m-1}$ and $\mu(d_m)=\xi_m$. Hence the unique optimal arm in round $m$ is
$d_m$, with $r^*(m)=\xi_m$, and the cumulative number of available arms is
$K(m)=|A(m)|=2m+1$. Choosing $N=10K(M)n_M$ yields
\(
\mathbb E R_T(\pi_{\mathrm{UCB\text{-}AA}})
=
\mathbb E R_M(\pi_{\mathrm{UCB\text{-}AA}})
=
O\bigl(\sqrt{T\log K(M)}\bigr).
\)
\end{example}

Example~\ref{exp:monotone-near-tied-ladder} should be interpreted as an endogenous-checkpoint
illustration of the same local-testing mechanism. The deterministic-checkpoint lower bound gives
the canonical minimax benchmark, while the UCB-AA analysis yields the same square-root-type
behavior up to an additional logarithmic factor along its own random checkpoint schedule.

Taken together, these lower-bound examples show that MAB-AA contains several
distinct sources of statistical difficulty. They are not intended as term-by-term
converses to the upper bounds in Section~\ref{sec:algorithm}; rather, they clarify
why the analysis must separately control the costs of identifying good incumbent
arms, certifying arriving competitors, and tracking benchmark changes.

\section{Simulation}\label{sec: simulation}
We evaluate UCB-AA on two synthetic arriving-arm environments. The first
contains abrupt improvements in the best available arm, while the second
contains frequent near-optimal arrivals. Both settings use horizon \(T=60000\).
We report three metrics:
\begin{itemize}
	    \item Cumulative dynamic regret \(R_t\): measures loss relative to the best
	    currently available arm,
	    \(R_t=\sum_{s=1}^t(r^*(s)-r_{I_s})\), where
	    \(r^*(s)=\max_{i\in\mathcal A_s}r_i\) and \(I_s\) is the arm pulled at
	    time \(s\).
    \item Wasted pulls \(W_T\): measures exploration spent on clearly suboptimal
    choices,
    \(W_T=\sum_{s=1}^T\mathbb I\{\Delta_{I_s}(s)\ge0.1\}\).
    \item Active-set size: measures the pruning effect of elimination, defined
    as the number of arms retained after elimination steps.
\end{itemize}

We consider two distinct arrival scenarios designed to stress-test different challenges inherent to the MAB-AA setting; for specific distribution parameter settings, refer to the supplementary materials: 
\subsection*{Scenario 1 (Progressive Step-up Arrivals)}
The environment starts with \(K_0=10\) arms and background arrivals following a
Poisson process with rate \(\lambda=1/200\), with means drawn from a bi-modal
mixture. To create benchmark shifts, high-performing step arms are introduced at
\(t/T\in\{0.3,0.6,0.9\}\), with
\(\mu_k=\min(1,r_0+k\Delta_{\mathrm{step}})\), \(r_0=0.6\), and
\(\Delta_{\mathrm{step}}=0.05\). This setting tests whether an algorithm can
quickly detect and switch to newly arrived superior arms.

\subsection*{Scenario 2 (Near-tied Top Set)}
The environment starts with \(K_0=20\) arms, including \(m_0=5\) near-optimal
arms with means from \(\mathrm{Uniform}(0.7,0.8)\). New arms arrive according
to a Bernoulli process with probability \(\lambda=1/200\), and \(30\%\) of them
are near-optimal. This setting represents mature decision problems with many
near-tied top arms, where small performance differences must be detected without
excessive exploration.

To ensure a fair comparison, we adapt classical algorithms to handle the continuous arrival of new arms. Their implementations are described below; additional details are provided in the supplementary materials.
\begin{enumerate}

     \item UCB-AA (Proposed): the proposed method, which buffers newly arrived
     arms and screens them through pre-elimination before full competition. We
     set \(N=e/\tilde{\Delta}_M^2\) in the simulations.
    \item Growing UCB1 (G-UCB1): UCB1 \parencite{Auer2002FinitetimeAO} with
    immediate admission of new arms, using index
    \(\hat{\mu}_{i,n}+\sqrt{2\ln t/n}\).
    \item Growing MOSS (G-MOSS): MOSS \parencite{audibert2009minimax} with
    immediate admission of new arms, using index
    \(\hat{\mu}_i+\sqrt{\max\{0,\log(T/(K_t n_i))\}/n_i}\), where
    \(K_t=|\mathcal A_t|\).

\item Dynamic Successive Elimination (D-SE): an SE-style baseline
\parencite{PerchetRigollet2013} that admits new arms at each round, samples them
to match incumbent pull counts, and then applies the SE elimination rule.

\item Naive Elimination (NE): UCB-AA with the same formal elimination rule but
without the pre-elimination stage.

\end{enumerate}

For each scenario, we ran 100 independent simulations. The figures below report
the average cumulative dynamic regret and wasted pulls, with 95\% confidence
intervals.

\begin{figure}[htbp]
    \centering
    \includegraphics[width=0.9\textwidth]{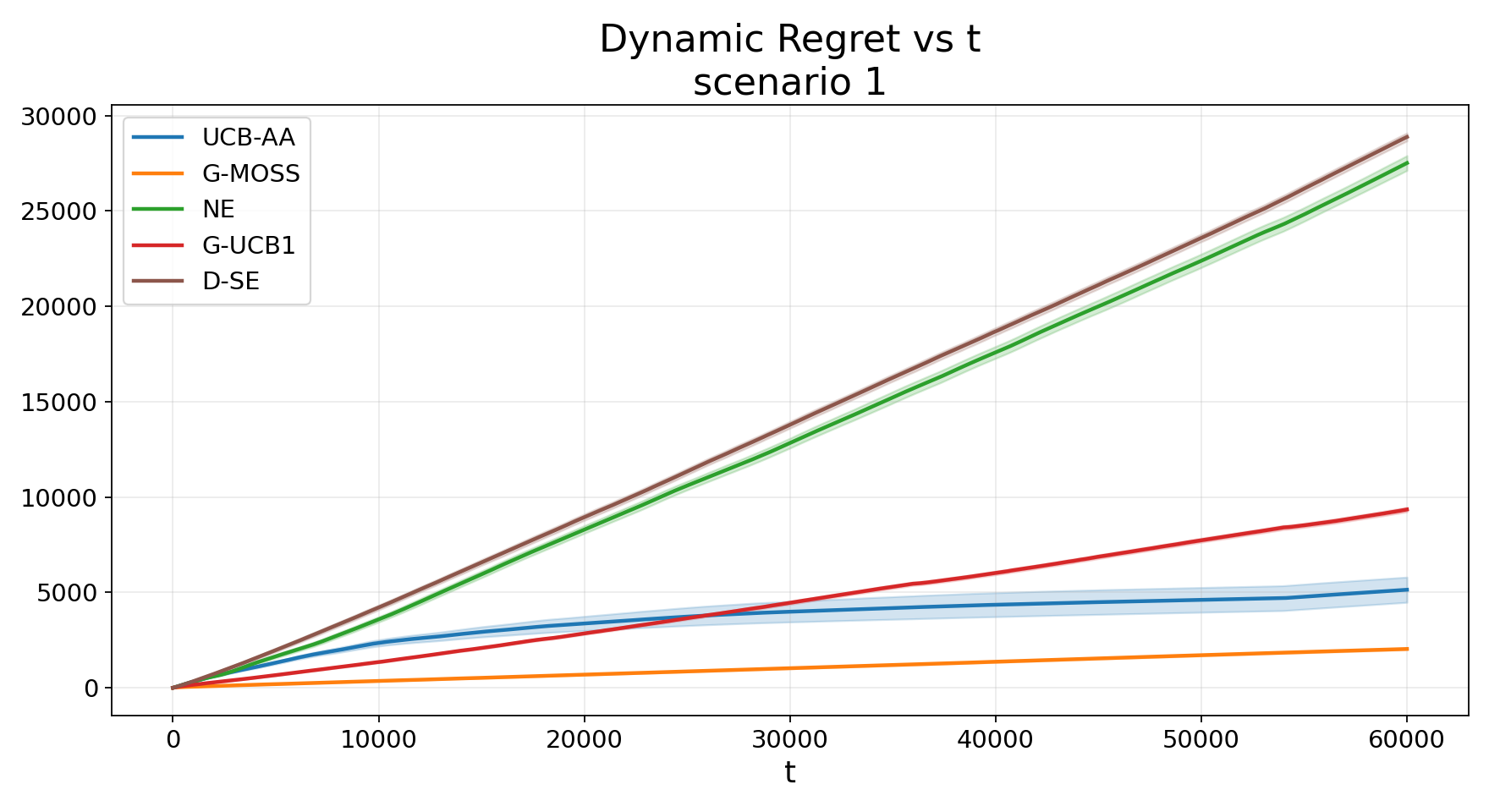}

    \smallskip
    \textit{(a) Cumulative dynamic regret (Scenario 1)}
    \vspace{8mm}

    \medskip
    \includegraphics[width=0.9\textwidth]{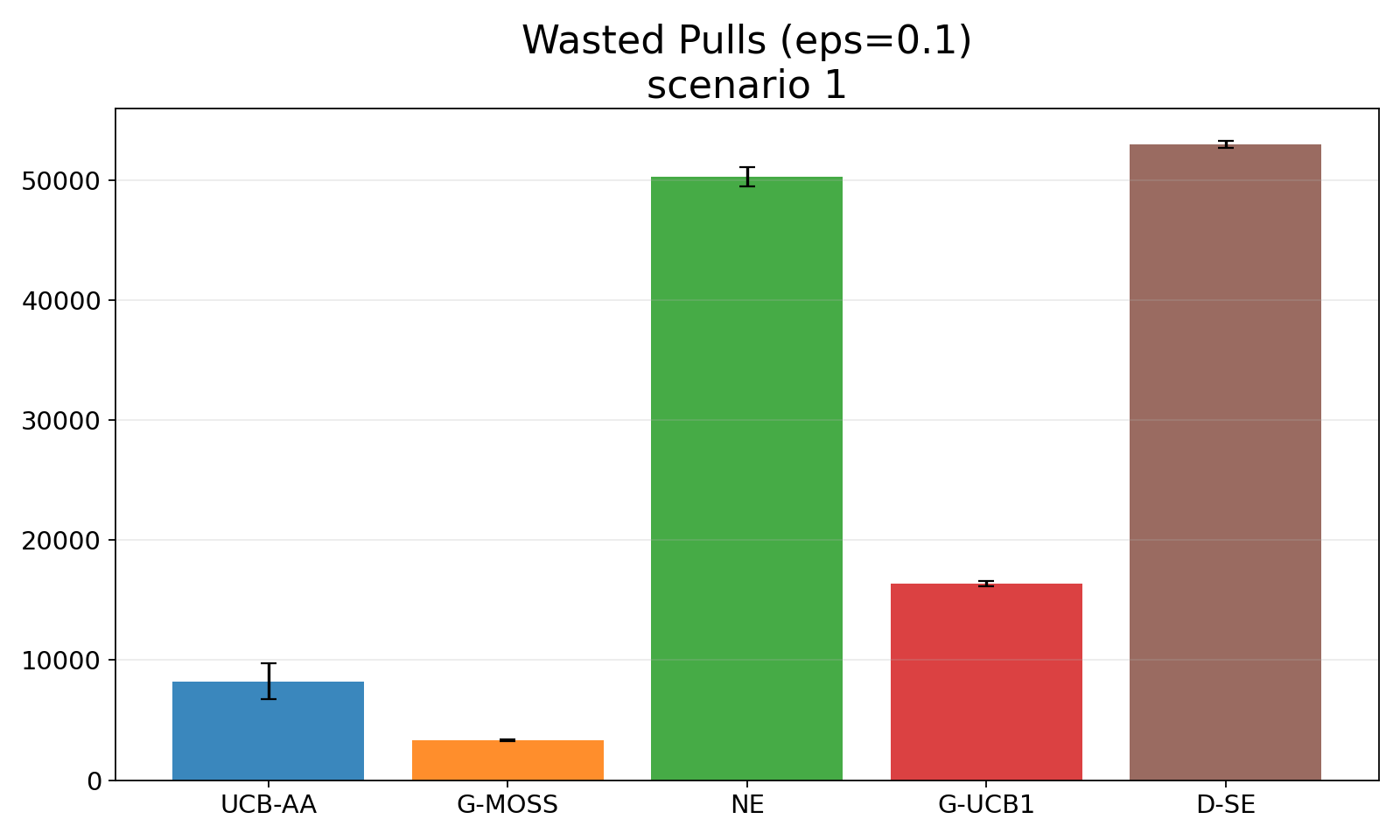}

    \smallskip
    \textit{(b) Wasted pulls (Scenario 1)}
    \caption{Experimental results for Scenario 1 (Progressive step-up arrivals). Panel (a) shows cumulative dynamic regret, and panel (b) compares wasted pulls ($W_T$) and the final number of active arms. Results are averaged over 100 independent runs.}
    \label{fig:scenario_a_results}
\end{figure}

\begin{figure}[htbp]
    \centering
    \includegraphics[width=0.9\textwidth]{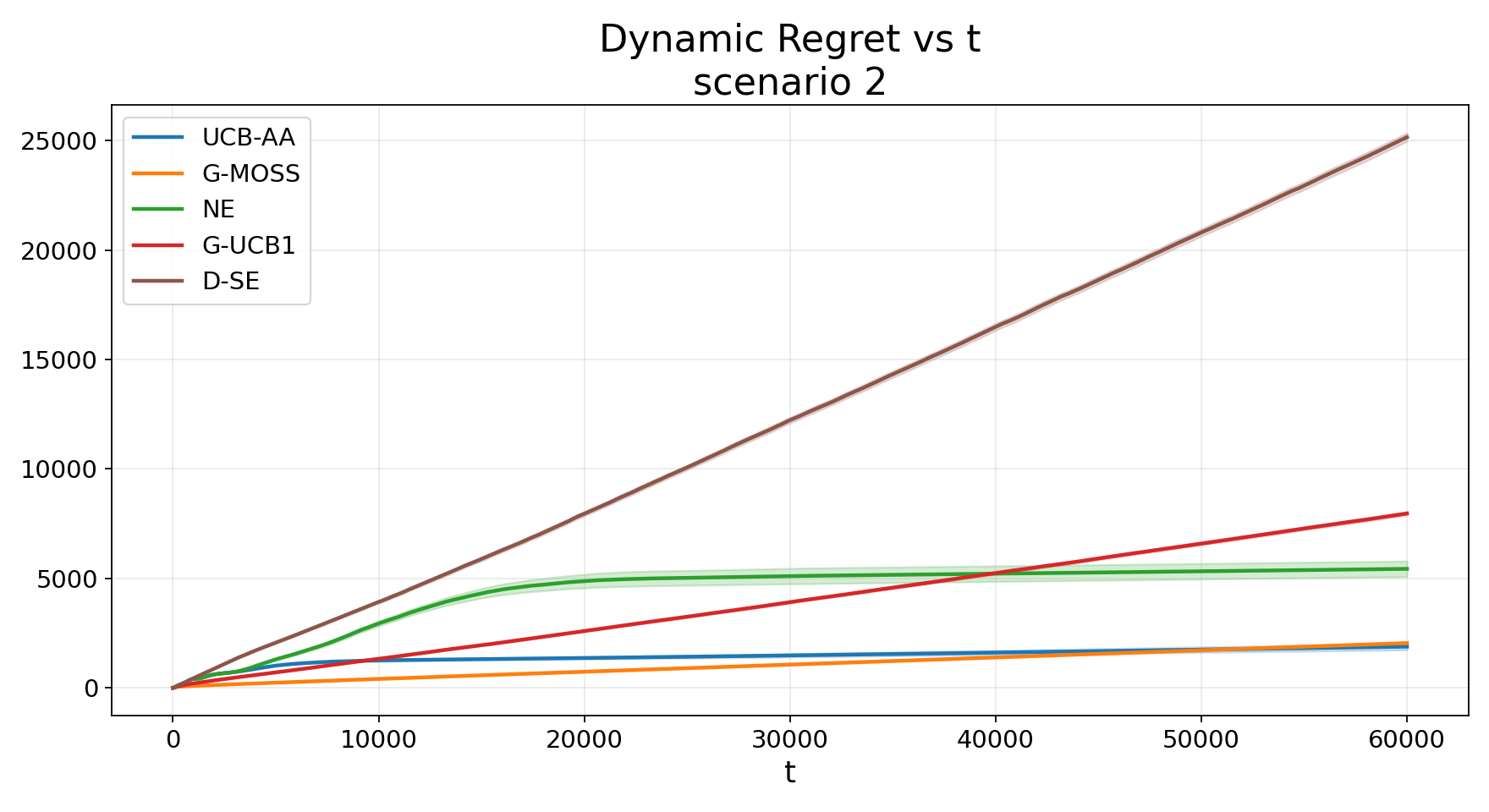}

    \smallskip
    \textit{(a) Cumulative dynamic regret (Scenario 2)}
    \vspace{8mm}

    \medskip
    \includegraphics[width=0.9\textwidth]{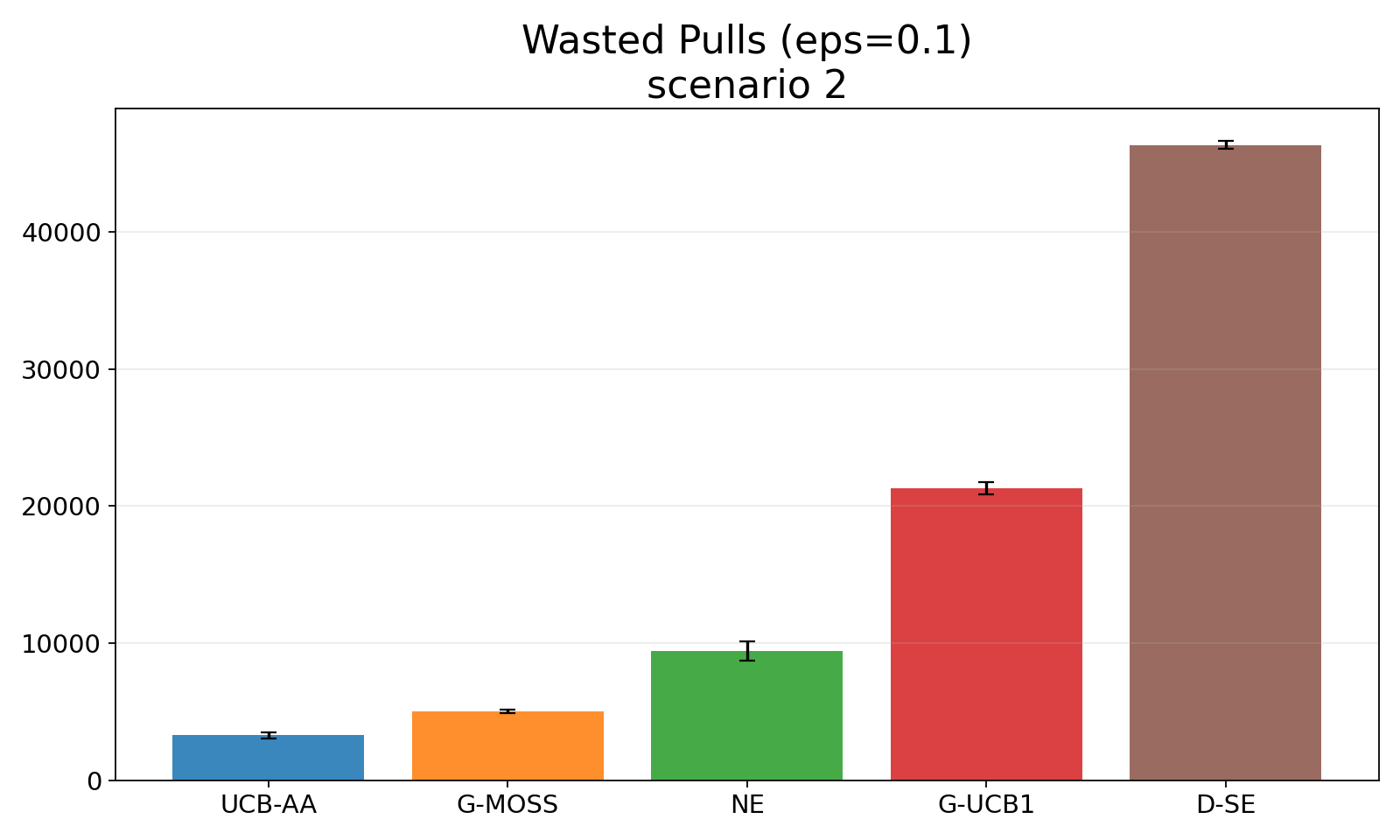}

    \smallskip
    \textit{(b) Wasted pulls (Scenario 2)}
    \caption{Experimental results for Scenario 2 (Near-tied top set). UCB-AA's explicit elimination mechanism maintains a compact decision space while achieving regret performance comparable to Growing MOSS.}

\end{figure}

\begin{figure}[htbp]
    \centering
    \includegraphics[width=0.9\textwidth]{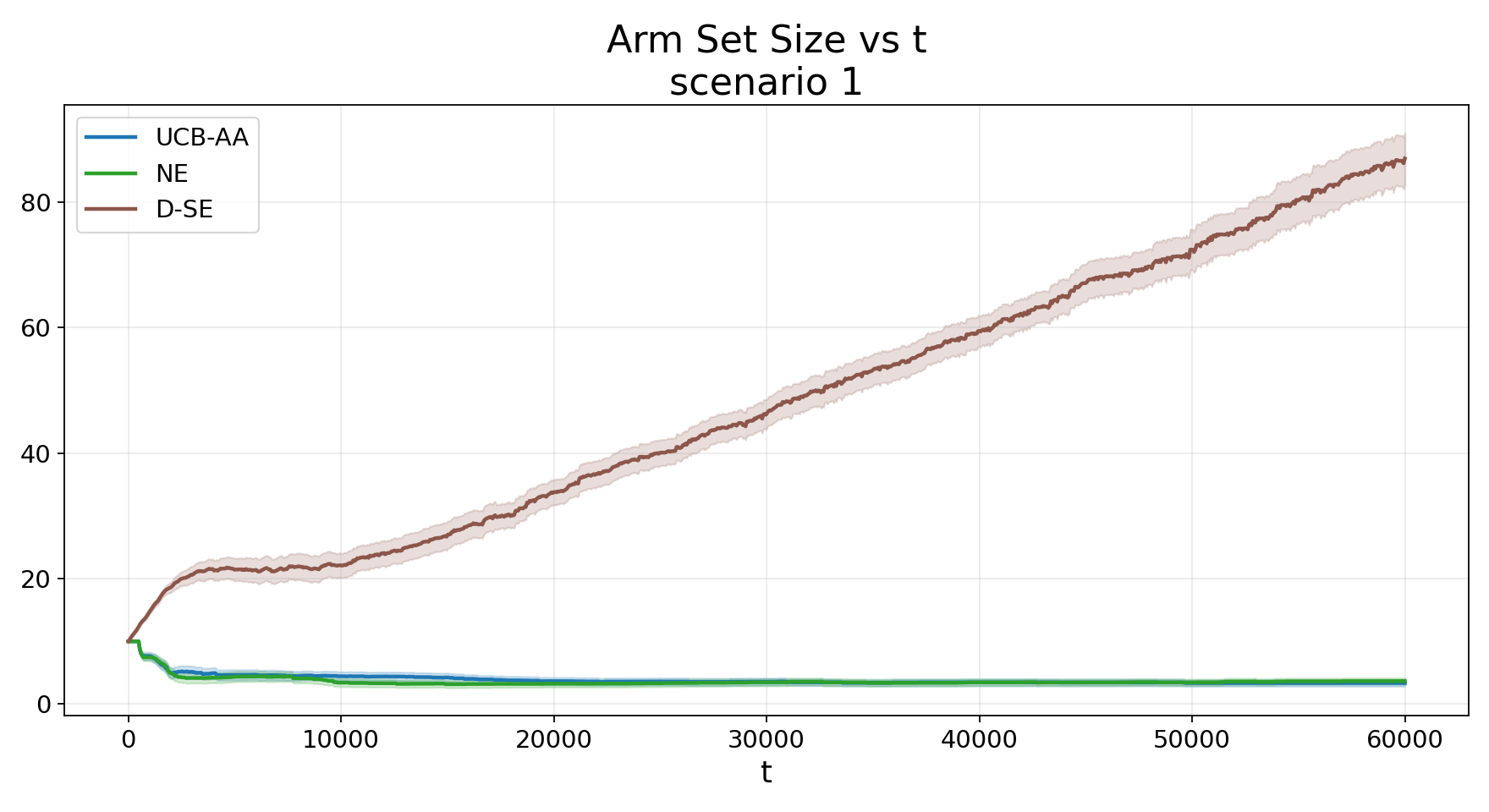}

    \smallskip
    \textit{(a) Active arm size (Scenario 1)}
    \vspace{8mm}

    \medskip
    \includegraphics[width=0.9\textwidth]
    {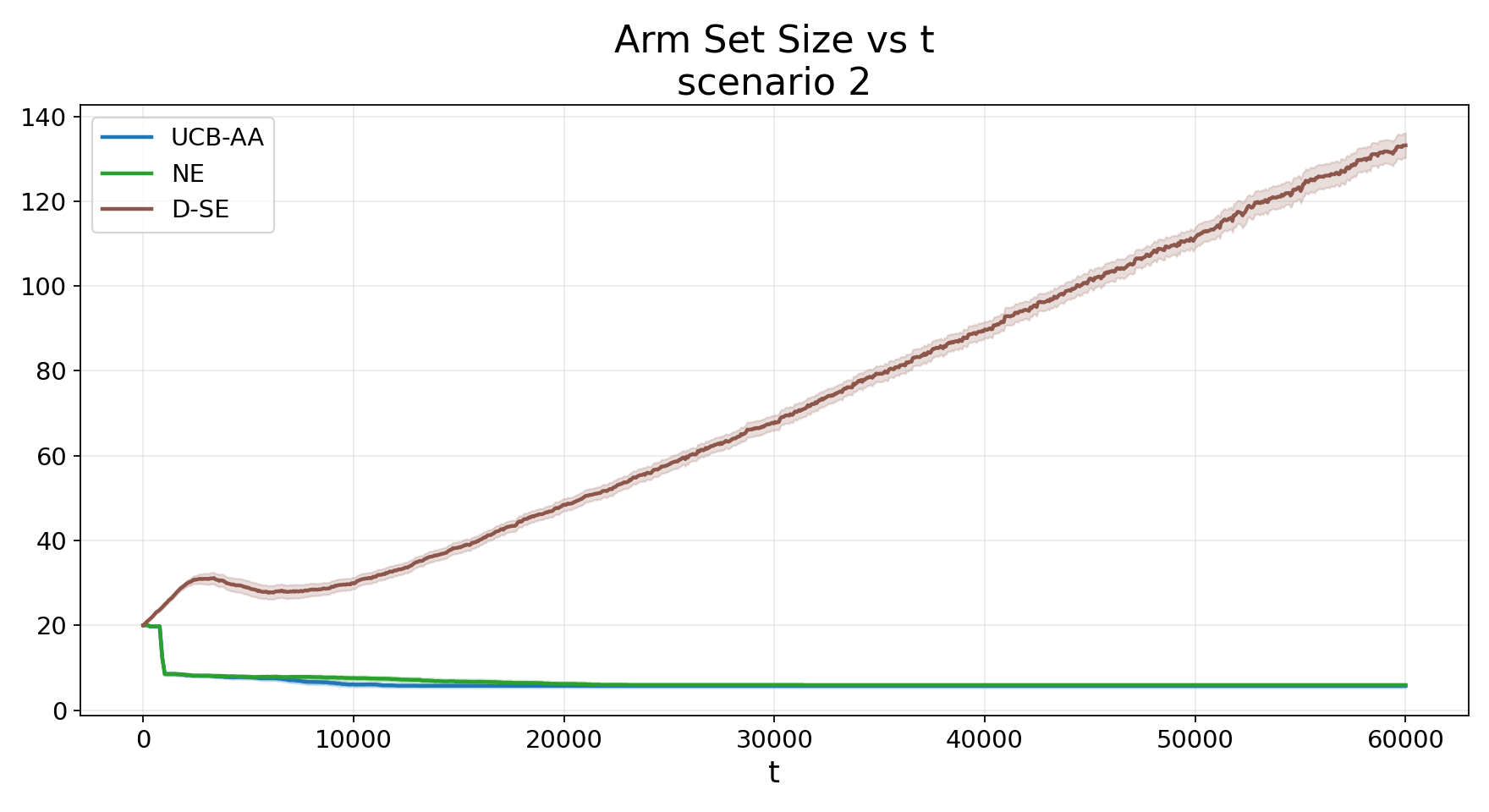}

    \smallskip
    \textit{(b) Active arm size (Scenario 2)}
    \caption{Active arm size for two scenarios. UCB-AA's explicit elimination mechanism maintains a compact decision space while achieving regret performance comparable to Growing MOSS.}
\end{figure}

The simulation results show that UCB-AA substantially reduces the number of wasted pulls
($W_T$) relative to Growing UCB1 and the elimination-based baselines. The main reason is that
newly arriving arms are not allowed to compete immediately on equal footing with well-sampled
incumbents. Instead, UCB-AA uses a buffered pre-elimination step to screen out clearly inferior
late arrivals before full competition, thereby protecting the sampling budget of the currently
most competitive arms. By contrast, Growing UCB1 continues to allocate exploration to each new
arrival through its uncertainty bonus, while Naive Elimination and Dynamic SE treat new and old
arms too symmetrically at entry, leading to larger regret.

A second advantage of UCB-AA is that it maintains a compact active set whose members have
survived repeated statistical comparisons over time. Methods such as Growing UCB1 and Growing MOSS do not explicitly prune the arm
set once arms arrive, whereas UCB-AA actively shrinks its working pool $|\mathcal{C}_t|$ through
pre-elimination and formal elimination. As a result, UCB-AA not only controls regret well, but
also returns a smaller set of arms that have survived repeated statistical comparisons and are
therefore more likely to be genuinely competitive.

At the same time, UCB-AA does not uniformly dominate Growing MOSS in cumulative dynamic regret
($R_T$); instead, their regret levels are often comparable, and Growing MOSS can be slightly
better in settings where many arms are near-tied. This reflects the cost of explicit hard
elimination. In UCB-AA, an arm is removed only when its upper confidence bound falls below
the largest lower confidence bound among the surviving arms,
\(
UCB_j < \max_i LCB_i.
\)
Because this rule requires strong evidence, UCB-AA may spend additional samples to certify that
a borderline arm is truly inferior before discarding it. By contrast, Growing MOSS performs only
a soft form of elimination: weak arms receive fewer pulls when their indices fall behind, but
they remain in the available set and can be sampled again if their uncertainty remains large.

This trade-off also indicates when hard elimination is useful. If the only objective is cumulative
regret over a moderate horizon and keeping all arrived arms active is essentially free, then soft
index-based methods such as Growing MOSS can be highly competitive. However, in many arriving-arm
applications, the active set itself is costly to maintain. Examples include recommendation systems
with continuously added items, adaptive A/B testing with many newly launched designs, biomedical
screening or dose-finding studies with costly or risky candidates, and sequential model or policy
selection where each surviving candidate requires repeated evaluation. In such settings, reducing
the active set is not merely a computational convenience; it is part of the statistical objective.
UCB-AA therefore trades a modest amount of additional exploration for certified pruning of clearly
inferior late arrivals, thereby controlling both wasted pulls and the operational burden created
by an expanding action space.

\section{Discussion}\label{sec:Discussion}

We studied stochastic multi-armed bandits with arriving arms, where the available action
set expands over time and the performance is measured against the best currently available arm. The distinct features of arrival information discrepancy (AID) and drifting benchmark (DB) depart from the classical fixed-oracle benchmark and creates two
coupled statistical challenges: repeated certification of newly arriving competitors and a moving target to catch up.

Our main contribution shows that these challenges can be handled in a structured way. The round-based reduction separates the principal certification burden from the additional delay induced by benchmark movement, while UCB-AA uses a pre-elimination stage to screen late arrivals before they enter full competition. Under suitable control of gap evolution, this yields sublinear regret together with a compact active set. The lower-bound regimes further clarify that the hardness of MAB-AA has multiple sources: the static regime recovers the classical logarithmic identification barrier, distractor arrivals induce cumulative certification costs, and near-tied improvements create a local-testing difficulty under a drifting benchmark.

Several extensions are natural. One direction is to incorporate contextual information or arm-level covariates when  available, which may improve reward via individualization as well as the screening efficiency for newly arriving arms. Another direction is to study settings with post-arrival reward drift, while retaining the central focus on how an expanding action set affects exploration and regret. Finally, the arriving-arm framework can be extended to settings involving delayed feedback, operational resource constraints, and arrival processes calibrated from real application data.

\printbibliography

@article{LAI19854,
title = {Asymptotically efficient adaptive allocation rules},
journal = {Advances in Applied Mathematics},
volume = {6},
% number = {1},
pages = {4-22},
year = {1985},
issn = {0196-8858},
doi = {https://doi.org/10.1016/0196-8858(85)90002-8},
url = {https://www.sciencedirect.com/science/article/pii/0196885885900028},
author = {T.L Lai and Herbert Robbins}
}

@article{qi2025graphfeedback,
      title={Graph Feedback Bandits on Similar Arms: With and Without Graph Structures}, 
      author={Han Qi and Fei Guo and Li Zhu and Qiaosheng Zhang},
      year={2025},
      journal = {arXiv preprint},
      eprint={2501.14314},
      archivePrefix={arXiv},
      primaryClass={cs.LG},
      url={https://arxiv.org/abs/2501.14314}, 
}

@article{Auer2002FinitetimeAO,
  author  = {Auer, Peter and Cesa-Bianchi, Nicol{\`o} and Fischer, Paul},
  title   = {Finite-time Analysis of the Multiarmed Bandit Problem},
  journal = {Machine Learning},
  volume  = {47},
  % number  = {2--3},
  pages   = {235--256},
  year    = {2002},
  doi     = {10.1023/A:1013689704352},
  url     = {https://doi.org/10.1023/A:1013689704352}
}

@article{Auer2010UCBRI,
      author = {Peter Auer and Ronald Ortner},
      title = {UCB revisited: Improved regret bounds for the stochastic multi-armed bandit problem},
      journal = {Periodica Mathematica Hungarica},
      year = {2010},
      publisher = {Akadémiai Kiadó, co-published with Springer Science+Business Media B.V., Formerly Kluwer Academic Publishers B.V.},
      address = {Budapest, Hungary},
      volume = {61},
      % number = {1-2},
      doi = {10.1007/s10998-010-3055-6},
      pages=      {55 - 65},
      url = {https://www.akjournals.com/view/journals/10998/61/1-2/article-p55.xml}
}

@article{10.1214/aos/1069362389,
author = {Donald A. Berry and Robert W. Chen and Alan Zame and David C. Heath and Larry A. Shepp},
title = {{Bandit problems with infinitely many arms}},
volume = {25},
journal = {The Annals of Statistics},
% number = {5},
publisher = {Institute of Mathematical Statistics},
pages = {2103 -- 2116},
keywords = {bandit problems, dynamic allocation of Bernoulli processes, Sequential experimentation, staying with a winner, switching with a loser},
year = {1997},
doi = {10.1214/aos/1069362389},
URL = {https://doi.org/10.1214/aos/1069362389}
}

@inproceedings{NIPS2008_788d9869,
author = {Chakrabarti, Deepayan and Kumar, Ravi and Radlinski, Filip and Upfal, Eli},
title = {Mortal multi-armed bandits},
year = {2008},
isbn = {9781605609492},
% publisher = {Curran Associates Inc.},
% address = {Red Hook, NY, USA},
abstract = {We formulate and study a new variant of the k-armed bandit problem, motivated by e-commerce applications. In our model, arms have (stochastic) lifetime after which they expire. In this setting an algorithm needs to continuously explore new arms, in contrast to the standard k-armed bandit model in which arms are available indefinitely and exploration is reduced once an optimal arm is identified with near-certainty. The main motivation for our setting is online-advertising, where ads have limited lifetime due to, for example, the nature of their content and their campaign budgets. An algorithm needs to choose among a large collection of ads, more than can be fully explored within the typical ad lifetime.We present an optimal algorithm for the state-aware (deterministic reward function) case, and build on this technique to obtain an algorithm for the state-oblivious (stochastic reward function) case. Empirical studies on various reward distributions, including one derived from a real-world ad serving application, show that the proposed algorithms significantly outperform the standard multi-armed bandit approaches applied to these settings.},
booktitle = {Proceedings of the 22nd International Conference on Neural Information Processing Systems},
pages = {273–280},
numpages = {8},
% location = {Vancouver, British Columbia, Canada},
series = {NIPS'08}
}

@article{10.1007/s10994-010-5178-7,
author = {Kleinberg, Robert and Niculescu-Mizil, Alexandru and Sharma, Yogeshwer},
title = {Regret bounds for sleeping experts and bandits},
year = {2010},
issue_date = {September 2010},
publisher = {Kluwer Academic Publishers},
address = {USA},
volume = {80},
% number = {2–3},
issn = {0885-6125},
url = {https://doi.org/10.1007/s10994-010-5178-7},
doi = {10.1007/s10994-010-5178-7},
abstract = {We study on-line decision problems where the set of actions that are available to the decision algorithm varies over time. With a few notable exceptions, such problems remained largely unaddressed in the literature, despite their applicability to a large number of practical problems. Departing from previous work on this "Sleeping Experts" problem, we compare algorithms against the payoff obtained by the best ordering of the actions, which is a natural benchmark for this type of problem. We study both the full-information (best expert) and partial-information (multi-armed bandit) settings and consider both stochastic and adversarial rewards models. For all settings we give algorithms achieving (almost) information-theoretically optimal regret bounds (up to a constant or a sub-logarithmic factor) with respect to the best-ordering benchmark.},
journal = {Machine Learning},
% month = sep,
pages = {245–272},
numpages = {28},
keywords = {Regret, Online algorithms, Computational learning theory}
}

@inproceedings{Besbes2014,
author = {Besbes, Omar and Gur, Yonatan and Zeevi, Assaf},
title = {Stochastic multi-armed-bandit problem with non-stationary rewards},
year = {2014},
% publisher = {MIT Press},
% address = {Cambridge, MA, USA},
abstract = {In a multi-armed bandit (MAB) problem a gambler needs to choose at each round of play one of K arms, each characterized by an unknown reward distribution. Reward realizations are only observed when an arm is selected, and the gambler's objective is to maximize his cumulative expected earnings over some given horizon of play T. To do this, the gambler needs to acquire information about arms (exploration) while simultaneously optimizing immediate rewards (exploitation); the price paid due to this trade off is often referred to as the regret, and the main question is how small can this price be as a function of the horizon length T. This problem has been studied extensively when the reward distributions do not change over time; an assumption that supports a sharp characterization of the regret, yet is often violated in practical settings. In this paper, we focus on a MAB formulation which allows for a broad range of temporal uncertainties in the rewards, while still maintaining mathematical tractability. We fully characterize the (regret) complexity of this class of MAB problems by establishing a direct link between the extent of allowable reward "variation" and the minimal achievable regret, and by establishing a connection between the adversarial and the stochastic MAB frameworks.},
booktitle = {Proceedings of the 28th International Conference on Neural Information Processing Systems},
volume = {1},
pages = {199–207},
numpages = {9},
% location = {Montreal, Canada},
series = {NIPS'14}
}

@article{ghalme2021ballooning,
title = {Ballooning multi-armed bandits},
journal = {Artificial Intelligence},
volume = {296},
pages = {103485},
year = {2021},
issn = {0004-3702},
doi = {https://doi.org/10.1016/j.artint.2021.103485},
url = {https://www.sciencedirect.com/science/article/pii/S0004370221000369},
author = {Ganesh Ghalme and Swapnil Dhamal and Shweta Jain and Sujit Gujar and Y. Narahari}
}

@inproceedings{audibert2009minimax,
  author = {Audibert, Jean-Yves and Bubeck, Sébastien},
title = {Minimax Policies for Adversarial and Stochastic Bandits},
booktitle = {Proceedings of the 22nd Annual Conference on Learning Theory (COLT)},
year = {2009},
pages   = {217--226},
 series = 	 {Proceedings of Machine Learning Research},
% month = jan,
abstract = {We fill in a long open gap in the characterization of the minimax rate for the multi-armed bandit problem. Concretely, we remove an extraneous logarithmic factor in the previously known upper bound and propose a new family of randomized algorithms based on an implicit normalization, as well as a new analysis. We also consider the stochastic case, and prove that an appropriate modification of the upper confidence bound policy UCB1 (Auer et al., 2002) achieves the distribution-free optimal rate while still having a distribution-dependent rate logarithmic in the number of plays.},
url = {https://www.microsoft.com/en-us/research/publication/minimax-policies-adversarial-stochastic-bandits/},
%edition = {Proceedings of the 22nd Annual Conference on Learning Theory (COLT)},
%note = {Best Student Paper Award},
}

@article{EvenDar2006SE,
  author  = {Eyal Even-Dar and Shie Mannor and Yishay Mansour},
  title   = {Action Elimination and Stopping Conditions for the Multi-Armed Bandit and Reinforcement Learning Problems},
  journal = {Journal of Machine Learning Research},
  year    = {2006},
  volume  = {7},
  % number  = {39},
  pages   = {1079--1105},
  url     = {http://jmlr.org/papers/v7/evendar06a.html}
}

@article{BurnetasKatehakis1996,
title = {Optimal Adaptive Policies for Sequential Allocation Problems},
journal = {Advances in Applied Mathematics},
volume = {17},
% number = {2},
pages = {122-142},
year = {1996},
issn = {0196-8858},
doi = {https://doi.org/10.1006/aama.1996.0007},
url = {https://www.sciencedirect.com/science/article/pii/S019688589690007X},
author = {Apostolos N. Burnetas and Michael N. Katehakis},
abstract = {Consider the problem of sequential sampling frommstatistical populations to maximize the expected sum of outcomes in the long run. Under suitable assumptions on the unknown parameters[formula], it is shown that there exists a classCRof adaptive policies with the following properties: (i) The expectednhorizon reward[formula]under any policy π0inCRis equal to[formula], asn→∞, where[formula]is the largest population mean and[formula]is a constant. (ii) Policies inCRare asymptotically optimal within a larger classCUFof “uniformly fast convergent” policies in the sense that[formula], for any π∈CUFand any[formula]such that[formula]. Policies inCRare specified via easily computable indices, defined as unique solutions to dual problems that arise naturally from the functional form of[formula]. In addition, the assumptions are verified for populations specified by nonparametric discrete univariate distributions with finite support. In the case of normal populations with unknown means and variances, we leave as an open problem the verification of one assumption.}
}

@article{KaufmannCappeGarivier2016,
author = {Kaufmann, Emilie and Capp\'{e}, Olivier and Garivier, Aur\'{e}lien},
title = {On the complexity of best-arm identification in multi-armed bandit models},
year = {2016},
issue_date = {January 2016},
publisher = {JMLR.org},
volume = {17},
% number = {1},
issn = {1532-4435},
abstract = {The stochastic multi-armed bandit model is a simple abstraction that has proven useful in many different contexts in statistics and machine learning. Whereas the achievable limit in terms of regret minimization is now well known, our aim is to contribute to a better understanding of the performance in terms of identifying the m best arms. We introduce generic notions of complexity for the two dominant frameworks considered in the literature: fixed-budget and fixed-confidence settings. In the fixed-confidence setting, we provide the first known distribution-dependent lower bound on the complexity that involves information-theoretic quantities and holds when m ≥ 1 under general assumptions. In the specific case of two armed-bandits, we derive refined lower bounds in both the fixedcon fidence and fixed-budget settings, along with matching algorithms for Gaussian and Bernoulli bandit models. These results show in particular that the complexity of the fixed-budget setting may be smaller than the complexity of the fixed-confidence setting, contradicting the familiar behavior observed when testing fully specified alternatives. In addition, we also provide improved sequential stopping rules that have guaranteed error probabilities and shorter average running times. The proofs rely on two technical results that are of independent interest: a deviation lemma for self-normalized sums (Lemma 7) and a novel change of measure inequality for bandit models (Lemma 1).},
journal = {Journal of Machine Learning Research},
% month = jan,
pages = {1–42},
numpages = {42},
keywords = {best-arm identification, information-theoretic divergences, multi-armed bandit, pure exploration, sequential testing}
}

@InProceedings{Auer2019ADSWITCH,
  title = 	 {Adaptively Tracking the Best Bandit Arm with an Unknown Number of Distribution Changes},
  author =       {Auer, Peter and Gajane, Pratik and Ortner, Ronald},
  booktitle = 	 {Proceedings of the 32nd Conference on Learning Theory},
  pages = 	 {138--158},
  year = 	 {2019},
  %editor = 	 {Beygelzimer, Alina and Hsu, Daniel},
  volume = 	 {99},
  series = 	 {Proceedings of Machine Learning Research},
 %  month = 	 jun,
 % publisher =    {PMLR},
  pdf = 	 {http://proceedings.mlr.press/v99/auer19a/auer19a.pdf},
  url = 	 {https://proceedings.mlr.press/v99/auer19a.html},
  abstract = 	 {We consider the variant of the stochastic multi-armed bandit problem where the stochastic reward distributions may change abruptly several times. In contrast to previous work, we are able to achieve (nearly) optimal mini-max regret bounds without knowing the number of changes. For this setting, we propose an algorithm called ADSWITCH and provide performance guarantees for the regret evaluated against the optimal non-stationary policy. Our regret bound is the first optimal bound for an algorithm that is not tuned with respect to the number of changes.}
}

@article{Woodroofe1979,
author = {Michael Woodroofe},
title = {A One-Armed Bandit Problem with a Concomitant Variable},
journal = {Journal of the American Statistical Association},
volume = {74},
% number = {368},
pages = {799--806},
year = {1979},
publisher = {Taylor \& Francis},
doi = {10.1080/01621459.1979.10481033},


URL = { 
    
    
        https://www.tandfonline.com/doi/abs/10.1080/01621459.1979.10481033
    

},
%eprint = {https://www.tandfonline.com/doi/pdf/10.1080/01621459.1979.10481033}

}

@article{YangZhu2002,
author = {Yuhong Yang and Dan Zhu},
title = {{Randomized Allocation with nonparametric estimation for a multi-armed bandit problem with covariates}},
volume = {30},
journal = {The Annals of Statistics},
% number = {1},
publisher = {Institute of Mathematical Statistics},
pages = {100 -- 121},
keywords = {con-comitant variable, Multi-armed bandits, Nonparametric regression, randomized allocation, Sequential allocation},
year = {2002},
doi = {10.1214/aos/1015362186},
URL = {https://doi.org/10.1214/aos/1015362186}
}

@inproceedings{RigolletZeevi2010,
title = "Nonparametric bandits with covariates",
abstract = "We consider a bandit problem which involves sequential sampling from two populations (arms). Each arm produces a noisy reward realization which depends on an observable random covariate. The goal is to maximize cumulative expected reward. We derive general lower bounds on the performance of any admissible policy, and develop an algorithm whose performance achieves the order of said lower bound up to logarithmic terms. This is done by decomposing the global problem into suitably {"}localized{"} bandit problems. Proofs blend ideas from nonparametric statistics and traditional methods used in the bandit literature.",
booktitle = {Proceedings of the 23rd Annual Conference on Learning Theory (COLT)},
 series = 	 {Proceedings of Machine Learning Research},
author = "Philippe Rigollet and Assaf Zeevi",
year = "2010",
isbn = "9780982252925",
% publisher = "Omnipress",
pages = "54--66",
% note = "23rd Conference on Learning Theory, COLT 2010 ; Conference date: 27-06-2010 Through 29-06-2010",
}

@article{PerchetRigollet2013,
author = {Vianney Perchet and Philippe Rigollet},
title = {{The multi-armed bandit problem with covariates}},
volume = {41},
journal = {The Annals of Statistics},
% number = {2},
publisher = {Institute of Mathematical Statistics},
pages = {693 -- 721},
keywords = {adaptive partition, contextual bandit, multi-armed bandit, Nonparametric bandit, regret bounds, Sequential allocation, successive elimination},
year = {2013},
doi = {10.1214/13-AOS1101},
URL = {https://doi.org/10.1214/13-AOS1101}
}

@article{QianYang2016,
author = {Wei Qian and Yuhong Yang},
title = {{Randomized allocation with arm elimination in a bandit problem with covariates}},
volume = {10},
journal = {Electronic Journal of Statistics},
% number = {1},
publisher = {Institute of Mathematical Statistics and Bernoulli Society},
pages = {242 -- 270},
keywords = {adaptive estimation, Contextual bandit problem, MABC, Nonparametric bandit, regret bound},
year = {2016},
doi = {10.1214/15-EJS1104},
URL = {https://doi.org/10.1214/15-EJS1104}
}

@article{VillarBowdenWason2015,
    author = {Villar, Sof{\'i}a S. and Wason, James and Bowden, Jack},
    title = {Response-Adaptive Randomization for Multi-arm Clinical Trials Using the Forward Looking Gittins Index Rule},
    journal = {Biometrics},
    volume = {71},
    % number = {4},
    pages = {969-978},
    year = {2015},
 %   month = {12},
    abstract = {The Gittins index provides a well established, computationally attractive, optimal solution to a class of resource allocation problems known collectively as the multi-arm bandit problem. Its development was originally motivated by the problem of optimal patient allocation in multi-arm clinical trials. However, it has never been used in practice, possibly for the following reasons: (1) it is fully sequential, i.e., the endpoint must be observable soon after treating a patient, reducing the medical settings to which it is applicable; (2) it is completely deterministic and thus removes randomization from the trial, which would naturally protect against various sources of bias. We propose a novel implementation of the Gittins index rule that overcomes these difficulties, trading off a small deviation from optimality for a fully randomized, adaptive group allocation procedure which offers substantial improvements in terms of patient benefit, especially relevant for small populations. We report the operating characteristics of our approach compared to existing methods of adaptive randomization using a recently published trial as motivation.},
    issn = {0006-341X},
    doi = {10.1111/biom.12337},
    url = {https://doi.org/10.1111/biom.12337},
    % eprint = {https://academic.oup.com/biometrics/article-pdf/71/4/969/55141851/biometrics_71_4_969.pdf},
}

@article{WilliamsonVillar2020,
    author = {Williamson, S. Faye and Villar, Sof{\'i}a S.},
    title = {A Response-Adaptive Randomization Procedure for Multi-Armed Clinical Trials with Normally Distributed Outcomes},
    journal = {Biometrics},
    volume = {76},
    % number = {1},
    pages = {197-209},
    year = {2020},
   % month = {03},
    abstract = {We propose a novel response-adaptive randomization procedure for multi-armed trials with continuous outcomes that are assumed to be normally distributed. Our proposed rule is non-myopic, and oriented toward a patient benefit objective, yet maintains computational feasibility. We derive our response-adaptive algorithm based on the Gittins index for the multi-armed bandit problem, as a modification of the method first introduced in Villar et al. (Biometrics, 71, pp. 969-978). The resulting procedure can be implemented under the assumption of both known or unknown variance. We illustrate the proposed procedure by simulations in the context of phase II cancer trials. Our results show that, in a multi-armed setting, there are efficiency and patient benefit gains of using a response-adaptive allocation procedure with a continuous endpoint instead of a binary one. These gains persist even if an anticipated low rate of missing data due to deaths, dropouts, or complete responses is imputed online through a procedure first introduced in this paper. Additionally, we discuss how there are response-adaptive designs that outperform the traditional equal randomized design both in terms of efficiency and patient benefit measures in the multi-armed trial context.},
    issn = {0006-341X},
    doi = {10.1111/biom.13119},
    url = {https://doi.org/10.1111/biom.13119}
    %eprint = {https://academic.oup.com/biometrics/article-pdf/76/1/197/54278096/biometrics_76_1_197.pdf},
}

@article{AzizKaufmannRiviere2021,
author = {Aziz, Maryam and Kaufmann, Emilie and Riviere, Marie-Karelle},
title = {On multi-armed bandit designs for dose-finding clinical trials},
year = {2021},
issue_date = {January 2021},
publisher = {JMLR.org},
volume = {22},
% number = {1},
issn = {1532-4435},
abstract = {We study the problem of finding the optimal dosage in early stage clinical trials through the multiarmed bandit lens. We advocate the use of the Thompson Sampling principle, a flexible algorithm that can accommodate different types of monotonicity assumptions on the toxicity and efficacy of the doses. For the simplest version of Thompson Sampling, based on a uniform prior distribution for each dose, we provide finite-time upper bounds on the number of sub-optimal dose selections, which is unprecedented for dose-finding algorithms. Through a large simulation study, we then show that variants of Thompson Sampling based on more sophisticated prior distributions outperform state-of-the-art dose identification algorithms in different types of dose-finding studies that occur in phase I or phase I/II trials.},
journal = {Journal of Machine Learning Research},
%month = jan,
articleno = {14},
pages   = {686--723},
keywords = {Bayesian methods, Thompson sampling, phase I/II clinical trials, phase I clinical trials, adaptive clinical trials, multi-armed bandits}
}

@article{Robbins1952,
author = {Herbert Robbins},
title = {{Some aspects of the sequential design of experiments}},
volume = {58},
journal = {Bulletin of the American Mathematical Society},
% number = {5},
publisher = {American Mathematical Society},
pages = {527 -- 535},
year = {1952},
}

@book{BerryFristedt1985,
  author = {Berry, Donald A. and Fristedt, Bert},
  title = {Bandit Problems: Sequential Allocation of Experiments},
  publisher = {Chapman and Hall},
  address = {London},
  year = {1985}
}

@article{Gittins1979,
    author = {Gittins, J. C.},
    title = {Bandit Processes and Dynamic Allocation Indices},
    journal = {Journal of the Royal Statistical Society: Series B (Methodological)},
    volume = {41},
    % number = {2},
    pages = {148-164},
    year = {1979},
 %   month = {01},
    abstract = {The paper aims to give a unified account of the central concepts in recent work on bandit processes and dynamic allocation indices; to show how these reduce some previously intractable problems to the problem of calculating such indices; and to describe how these calculations may be carried out. Applications to stochastic scheduling, sequential clinical trials and a class of search problems are discussed.},
    issn = {0035-9246},
    doi = {10.1111/j.2517-6161.1979.tb01068.x},
    url = {https://doi.org/10.1111/j.2517-6161.1979.tb01068.x},
    %eprint = {https://academic.oup.com/jrsssb/article-pdf/41/2/148/49097474/jrsssb_41_2_148.pdf},
}

@article{QianIngLiu2023,
author = {Wei Qian and Ching-Kang Ing and Ji Liu},
title = {Adaptive Algorithm for Multi-Armed Bandit Problem with High-Dimensional Covariates},
journal = {Journal of the American Statistical Association},
volume = {119},
% number = {546},
pages = {970--982},
year = {2024},
publisher = {Taylor \& Francis},
doi = {10.1080/01621459.2022.2152343},

URL = { https://doi.org/10.1080/01621459.2022.2152343
    
    

},
%eprint = { https://doi.org/10.1080/01621459.2022.2152343}
}

@article{Lai1987,
author = {Tze Leung Lai},
title = {{Adaptive Treatment Allocation and the Multi-Armed Bandit Problem}},
volume = {15},
journal = {The Annals of Statistics},
% number = {3},
publisher = {Institute of Mathematical Statistics},
pages = {1091 -- 1114},
keywords = {adaptive control, boundary crossings, Dynamic allocation, Sequential experimentation, upper confidence bounds},
year = {1987},
doi = {10.1214/aos/1176350495},
URL = {https://doi.org/10.1214/aos/1176350495}
}

@article{Zelen1969,
author = {M. Zelen},
title = {Play the Winner Rule and the Controlled Clinical Trial},
journal = {Journal of the American Statistical Association},
volume = {64},
% number = {325},
pages = {131--146},
year = {1969},
publisher = {Taylor \& Francis},
doi = {10.1080/01621459.1969.10500959},


URL = { 
    
    
        https://www.tandfonline.com/doi/abs/10.1080/01621459.1969.10500959
    

},
%eprint = {https://www.tandfonline.com/doi/pdf/10.1080/01621459.1969.10500959}

}

@article{WeiDurham1978,
author = {L. J. Wei and S. Durham},
title = {The Randomized Play-the-Winner Rule in Medical Trials},
journal = {Journal of the American Statistical Association},
volume = {73},
% number = {364},
pages = {840--843},
year = {1978},
publisher = {Taylor \& Francis},
doi = {10.1080/01621459.1978.10480109},


URL = { 
    
    
        https://www.tandfonline.com/doi/abs/10.1080/01621459.1978.10480109
    

},
%eprint ={https://www.tandfonline.com/doi/pdf/10.1080/01621459.1978.10480109}

}

@article{CaiCaiLi2024,
author = {Changxiao Cai and T. Tony Cai and Hongzhe Li},
title = {{Transfer learning for contextual multi-armed bandits}},
volume = {52},
journal = {The Annals of Statistics},
% number = {1},
publisher = {Institute of Mathematical Statistics},
pages = {207 -- 232},
keywords = {Adaptivity, Contextual multi-armed bandit, covariate shift, Minimax rate, regret bounds, self-similarity, transfer learning},
year = {2024},
doi = {10.1214/23-AOS2341},
URL = {https://doi.org/10.1214/23-AOS2341}
}

\end{document}

\typeout{get arXiv to do 4 passes: Label(s) may have changed. Rerun}